\ifdefined\XeTeXversion\else\pdfoutput=1\fi

\documentclass[letterpaper, 10 pt, conference]{ieeeconf}  %

\IEEEoverridecommandlockouts                              %

\usepackage[OT1]{fontenc}
\usepackage{times}
\usepackage{cite}     
\usepackage{xspace}
\usepackage{float}
\usepackage{subcaption}
\usepackage{graphicx}
\usepackage{booktabs,balance,microtype}
\usepackage{multirow}
\usepackage{wrapfig}
\usepackage[table]{xcolor}
\usepackage[hidelinks]{hyperref}
\usepackage[nameinlink,capitalize]{cleveref}
\usepackage[detect-weight,detect-family]{siunitx}
\usepackage{makecell}
\usepackage{pifont}
\let\labelindent\relax %
\usepackage{enumitem}
\usepackage{cuted}

\newcommand{\algabbr}{RoboRender\xspace}
\makeatletter
\newcommand{\myparagraph}[1]{%
  \par
  \if@nobreak\else\vspace{-\parskip}\vspace{6pt}\fi
  \noindent\textbf{#1}\hspace{0.4em}\ignorespaces}
\makeatother

\title{\LARGE \bf
RoboRender: Robot-Oriented Video Generation for Visual \\ Sim-to-Real Transfer
}

\author{Huang Huang$^{1*}$, Wensi Ai$^{1*}$, Ziyu Chen$^{1*}$, Youhui Wang$^{1*}$, \\ Zijian Du$^{2}$, Yang Liu$^{3}$, Jiaolong Yang$^{3}$, Li Fei-Fei$^{1}$, Jiajun Wu$^{1}$\\[0.3em]
$^{1}$Stanford University  \qquad $^{2}$Nvidia \qquad $^{3}$Microsoft Research \qquad $^{*}$Equal Contribution
}

\begin{document}
\bstctlcite{IEEEexample:BSTcontrol}
\maketitle
\thispagestyle{empty}
\pagestyle{empty}

\begin{strip}
    \vspace*{-2.5em}
    \centering
    \setlength{\abovecaptionskip}{5pt}
        \includegraphics[width=\textwidth, clip]{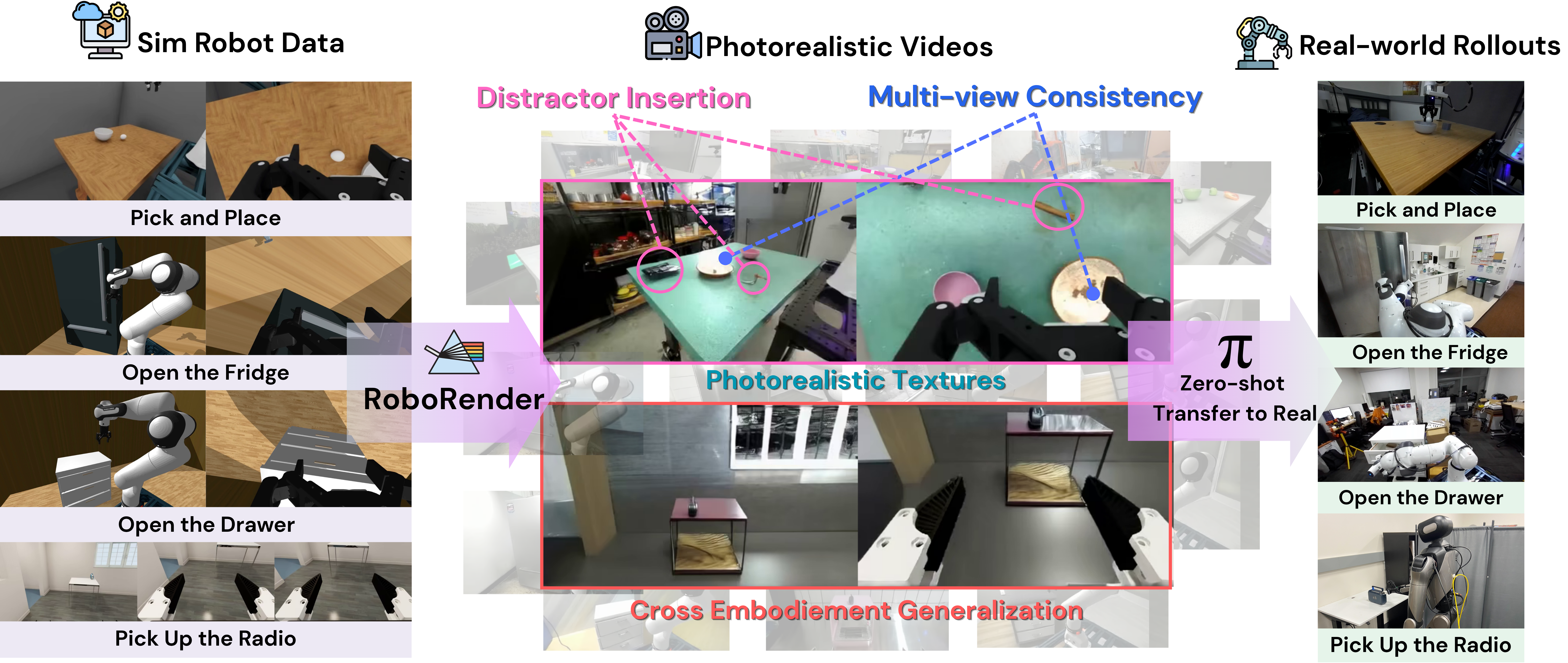}
  \captionof{figure}{\textbf{Overview.} From simulation trajectories (\textbf{Left}), \algabbr{} generates \textbf{diverse} and \textbf{photorealistic} videos oriented for visual sim-to-real robot policy transfer (\textbf{Middle}). These videos preserve simulator geometry and robot motion while supporting \textbf{multi-view consistency} across different robot camera views, \textbf{distractor insertion} for robustness, and \textbf{cross-embodiment generation} for unseen robot embodiments. Policies trained on \algabbr{}-generated videos achieve strong zero-shot sim-to-real performance across pick-and-place, articulated-object manipulation, and mobile manipulation tasks (\textbf{Right}), demonstrating that \algabbr{} effectively mitigates the visual sim-to-real gap.\looseness=-1}
    \label{fig:teaser}
    \vspace{-1.5em}
\end{strip}

\begin{abstract}

Simulation enables large-scale, low-cost robot data generation, but policies trained in simulation often fail to transfer to the real world due to the sim-to-real visual discrepancies. Existing approaches often rely on intermediate representations, which can discard rich semantic information or require additional perception modules at deployment. We address this visual sim-to-real gap with \algabbr{}, a framework that converts simulated trajectories into photorealistic RGB videos for policy learning. \algabbr{} trains a robot-oriented video generation model conditioned on simulated depth videos, language instructions, and robot RGB mask videos, preserving simulator geometry, robot motion, and action labels while synthesizing realistic textures, backgrounds, and distractors. The generated RGB videos are paired with simulator-provided states and actions to train policies for zero-shot real-world deployment. On robot video test sets, our video model outperforms depth-conditioned video generation baselines in generation quality. In real-world experiments across pick-and-place, articulated-object manipulation, and mobile manipulation tasks, policies trained on \algabbr{}-generated data achieve a 71\% average success rate, outperforming raw simulation renderings and conventional visual domain randomization by approximately \(7.1\times\) and \(3.6\times\), respectively. 
We further show that policy performance improves with more generated videos per simulation trajectory, increasing opening-task success by 65 percentage points. These results demonstrate that generative video rendering mitigates the visual sim-to-real gap for zero-shot policy transfer. Project website: \url{https://robo-render.github.io/}.
\end{abstract}

\section{Introduction}
Simulation provides privileged state information, supports large-scale parallel data generation, and enables low-cost and automatic environment reset, making it effective for robot policy learning. However, transferring simulation-trained policies to the real world remains challenging due to the sim-to-real gap, which includes both dynamics and visual discrepancies. For vision-based manipulation, the visual gap is particularly limiting, as policies must act from observations whose appearance can differ substantially between rendered and real scenes.  
Existing approaches reduce this gap by reconstructing realistic digital twins~\cite{kerbl20233d,torne2024robot,yu2025real2render2real,dan2025x}, replacing RGB with depth or object states~\cite{yuan2025hermes,lum2025crossing,kedia2026simtoolreal}, or applying visual domain randomization~\cite{he2025viral,deshpande2026molmob0t}, but can be scene-specific, semantically lossy, require additional perception modules, or remain visually mismatched to real observations.

Video generation models encode rich priors over real-world appearance, semantics, and temporal dynamics. In light of this, we propose \textbf{\algabbr{}}~(\cref{fig:teaser}), a framework that renders simulated trajectories into photorealistic RGB-action data for zero-shot visual sim-to-real policy transfer. At its core is a robot-oriented video generation model conditioned on simulated depth videos, language instructions, and robot RGB masks. Depth preserves scene geometry and object motion across views, enabling multi-view-consistent generation; language controls appearance, materials, lighting, and background semantics; and robot masks preserve robot motion and support generation for unseen embodiments. Together, these conditions allow \algabbr{} to synthesize rich, diverse and realistic textures, backgrounds, and distractors while retaining simulator geometry and action labels, making it efficient for generating visually diverse and physically grounded data. Policies are then trained on the resulting photorealistic RGB-action trajectories and deployed zero-shot in the real world.

We evaluate \algabbr{} on five pick-and-place tasks, two opening tasks, and one mobile manipulation task. Compared with policies trained on raw simulator RGB and with visual domain randomization, policies trained on \algabbr{}-generated data achieve \(\textbf{7.1}\times\) and \(\textbf{3.6}\times\) higher real-world success rates, respectively. We further show that generating multiple visually diverse videos from each simulated trajectory with different language prompts improves performance by up to \textbf{65} percentage points on opening tasks and \textbf{10} points on pick-and-place tasks. These results demonstrate the effectiveness of \algabbr{} in mitigating the visual sim-to-real gap for policy transfer.

In summary, our contributions are:
\begin{enumerate}[leftmargin=*]
    \item[-] We introduce \algabbr{}, a framework that converts simulated trajectories into photorealistic RGB demonstrations for zero-shot visual sim-to-real policy transfer.
    \item[-] We develop a robot-oriented video generation model conditioned on depth, language, and robot masks for grounded, controllable, and embodiment-aware video synthesis.
    \item[-] Real-robot experiments indicate that \algabbr{} enables significantly stronger zero-shot sim-to-real transfer than policies trained on raw simulation or conventional visual domain randomization, across pick-and-place, articulated-object manipulation, and mobile manipulation tasks.
\end{enumerate}

\section{Related Work}

\subsection{Visual Sim-to-Real Transfer}
Simulation enables scalable robot data generation, but policies trained in simulation often fail to transfer due to dynamics and visual discrepancies. While dynamics gaps can be mitigated through system identification, dynamics randomization, or adaptation~\cite{peng2018simtoreal,muratore2022randomizedsimulations}, vision-based manipulation remains particularly sensitive to the visual gap between simulation and real observations.
Prior work addresses this gap through three main strategies. Real2sim2real methods reconstruct photorealistic digital twins using neural rendering or 3D Gaussian Splatting~\cite{kerbl20233d,torne2024robot,yu2025real2render2real,dan2025x}, but require per-scene capture, calibration, or asset preparation, and may struggle with accurate interactive geometry~\cite{torne2024robot,yu2025real2render2real}. Geometry- or object-centric methods avoid sim-to-real RGB gap by using depth, point clouds, object poses, or object states~\cite{yuan2025hermes,qin2023dexpoint,lum2025crossing,kedia2026simtoolreal}, but can discard semantic appearance cues and require additional perception modules. Visual domain randomization trains policies on randomized textures, lighting, cameras, and backgrounds~\cite{tobin2017domainrandomization,he2025viral,deshpande2026molmob0t}, but often requires careful tuning and may remain far from real observations. 
In contrast, \algabbr{} retains the simulator as the source of geometry, dynamics, and action labels, while using a generative video prior to render simulated depth trajectories into photorealistic RGB videos. This approach combines the scalability and physical grounding of simulation with the rich semantic and visual priors of video generation models.

\subsection{Video Generation for Robot Data Synthesis}

Video generation models have emerged as synthetic data engines for robot learning. DreamGen~\cite{dreamgen2025} generates robot videos from language-conditioned world models, but language alone provides limited physical grounding, and actions must be inferred through inverse dynamics or latent action models. RoboCurate~\cite{kim2026robocurate} improves generation quality through simulator-replay filtering; \algabbr{} instead starts from simulator rollouts with known actions, preserving grounded dynamics and action labels. Other works add grounding through robot motion or simulator-derived cues: AnchorDream~\cite{ye2025anchordreamrepurposingvideodiffusion} conditions on language and robot motion videos but lacks grounded robot-object interactions, while EmbodieDreamer~\cite{wang2025embodiedreameradvancingreal2sim2realtransfer} and CRAFT~\cite{chen2026craftvideodiffusionbimanual} translate simulated observations using object masks or Canny edges, which provide 2D structure but lack explicit 3D information and multi-view consistency. \algabbr{} instead conditions on multi-view depth videos, preserving 3D scene structure for physically grounded, multi-view-consistent generation.
RoboTransfer~\cite{liu2026robotransfercontrollablegeometryconsistentvideo} also generates multi-view-consistent videos using depth, surface normals, and appearance exemplars such as background and object images, but reliance on pre-collected photorealistic conditions limits flexibility. \algabbr{} uses language to control textures, materials, lighting, and background semantics, enabling diverse synthesis without appearance exemplars. Compared with GigaWorld-0~\cite{gigaworld2025}, which studies video-model-based robot data generation across appearance, view, human-to-robot, and sim-to-real transfer, \algabbr{} focuses on visual sim-to-real transfer from simulated depth trajectories and supports distractor insertion and unseen-embodiment generalization through robot-mask conditioning.
Overall, \algabbr{} complements prior work with capabilities specialized for robot learning: physically grounded dynamics, simulator-provided action labels, multi-view consistency, distractor insertion, and embodiment generalization, helping mitigate the visual sim-to-real gap.

\section{Method}
\begin{figure*}[t]
    \centering
    \includegraphics[width=\linewidth]{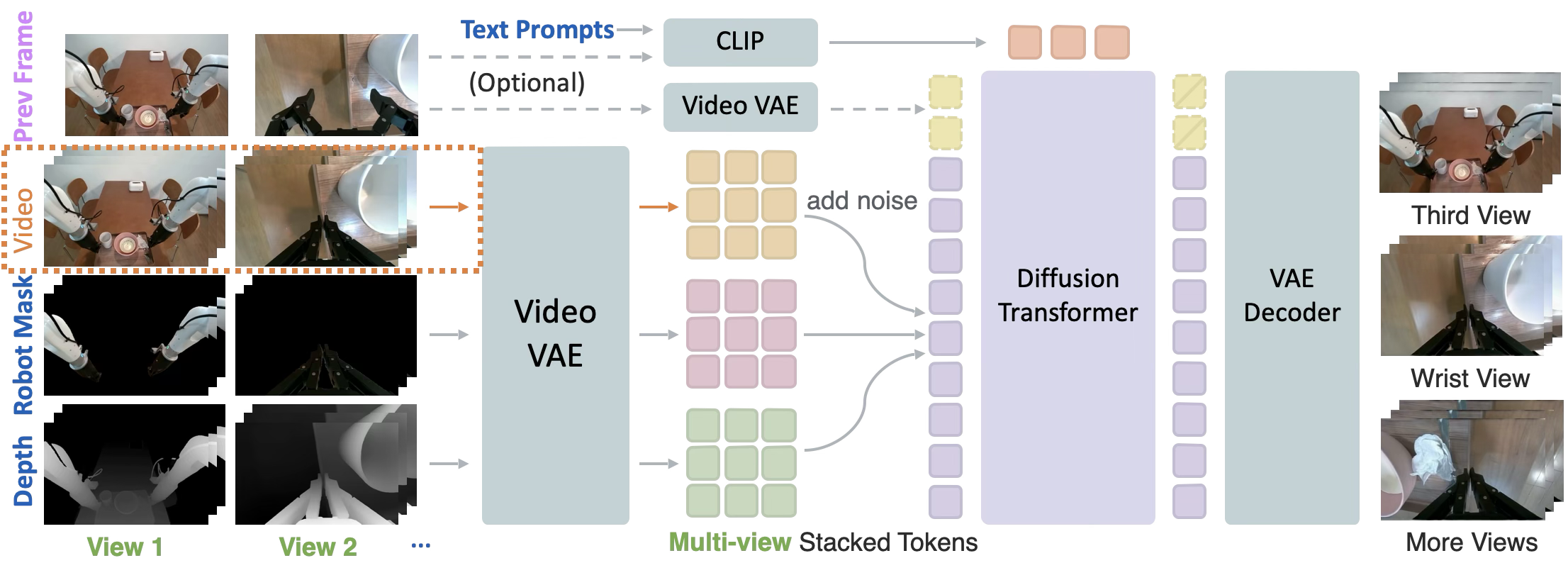}
    \vspace{-3pt}
    \caption{\textbf{RoboRender video generation model.} During training, raw RGB videos are VAE-encoded and noised to form the diffusion input. Per-view depth videos and robot-arm RGB mask videos are encoded as spatial conditions, while CLIP-encoded language prompts and previous-frame inputs provide semantic and temporal conditioning. The per-view condition latents are concatenated with the noised latents and stacked into a unified multi-view latent, which is denoised and decoded into RGB videos. At inference, ground-truth RGB videos (orange) are omitted and the diffusion transformer samples from noise conditioned on depth, robot-arm RGB, language, and previous frames.}
    \label{fig:video_model_pipeline}
    \vspace{-8pt}
\end{figure*}
\subsection{Video Generation Model}\label{ssec:video_gen}
\myparagraph{Model Architectures.}
RGB videos for robot learning must be photorealistic, physically grounded, cross-view consistent, and action-aligned. Existing text-to-video models~\cite{wan, cosmos} provide strong visual priors but often fail to capture accurate causal interactions~\cite{sora}. Depth-conditioned variants~\cite{wanfuncontrol, cosmostransfer} improve geometric control, yet usually generate views independently and lack robot-specific manipulation priors. We therefore adapt a variant of Wan2.1-T2V-1.3B~\cite{wan} into a robot-oriented video generation model for robot visual sim-to-real transfer. As shown in Fig.~\ref{fig:video_model_pipeline}, the model is conditioned on depth videos for scene geometry and object motion, robot-mask RGB videos for embodiment and robot motion, language prompts for appearance and scene semantics, and any available previous frame for autoregressive long-horizon generation. During training, the depth and robot-mask videos are VAE-encoded and channel-wise concatenated with the noised latent of the ground-truth RGB video, then stacked across views into a unified multi-view latent. A diffusion transformer then denoises this latent to synthesize photorealistic, multi-view-consistent videos while preserving simulator-provided structure and action labels for policy learning. We fine-tune the model with LoRA at a global batch size of $16$ for $12$K iterations on ${\sim}130$K real-world data.
Wan2.1-T2V-1.3B is a bidirectional diffusion model with a fixed 81-frame window, whereas robot demonstrations are variable-length and often longer. For long-horizon synthesis, we condition the model on the previous frame (purple in~\cref{fig:video_model_pipeline}) during training~\cite{skyreels, cosmos}. At inference, it generates autoregressively, conditioning each chunk on the previous chunk's last frame. We randomly drop this condition during training, enabling first-chunk synthesis without a previous frame and autoregressive continuation with one. During training, we additionally mask large random depth patches in each view, so the model learns to synthesize plausible, prompt-guided appearance where depth is absent rather than merely reproducing the input geometry. At inference, this enables a controllable scene edit: by dropping depth over a target region across all views and prompting for a new object, the model inserts a distractor that remains consistent across views. Used as data augmentation, these distractors make the downstream policy more robust to object clutter. We show examples of these inserted distractors in \cref{fig:teaser}.

\myparagraph{Training Data Curation.}
Video-model training requires large-scale paired conditioning signals and target videos. However, existing robotics datasets~\cite{agibot, droid} are primarily designed for policy learning, providing RGB observations and actions but not the conditions required by our model. We therefore build a scalable preprocessing pipeline that converts real-robot datasets into training-ready samples. For each clip, we extract a temporally aligned 81-frame window across all available views and generate four aligned conditions: a scene-description prompt, per-view depth videos, per-view robot-arm RGB masks, and a pre-window frame for temporal anchoring.
Given each real-robot clip, we generate scene prompts with Qwen2-VL~\cite{qwenvl} from the first frame of each view, estimate depth using Video-Depth-Anything~\cite{vda} and FoundationStereo~\cite{foundationstereo} when stereo is available, and extract robot-arm masks with SAM3~\cite{sam3}. Further preprocessing details are provided in Appendix~\ref{app:data_curation}.
Our final video-model training set contains ${\sim}130\mathrm{k}$ real-world robot manipulation samples: ${\sim}70\mathrm{k}$ from AgiBot-World-Alpha~\cite{agibot} and ${\sim}60\mathrm{k}$ from DROID~\cite{droid}, processed at $416{\times}240$ resolution. 
\begin{figure*}[t]
    \centering
    \includegraphics[width=\linewidth]{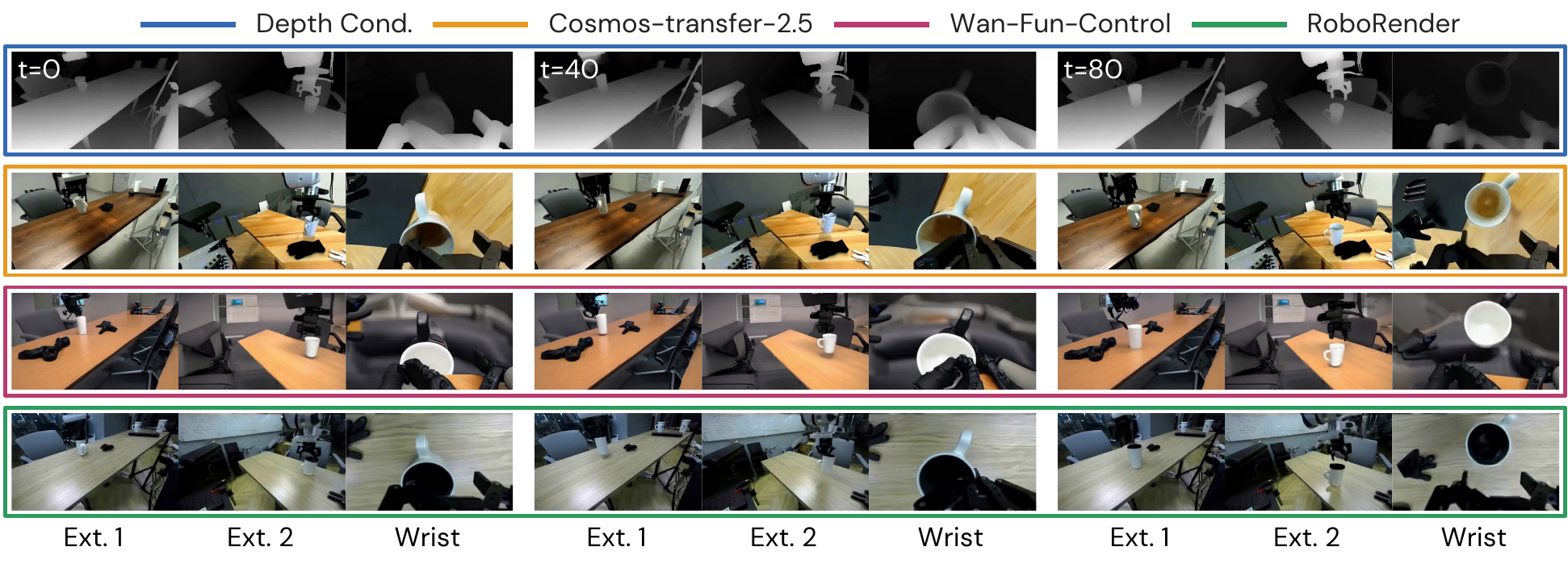}
    \vspace{-10pt}
    \caption{\textbf{Visual comparison of video generation.} Given depth conditions, we compare RGB videos generated by \algabbr{} against the baselines Cosmos-Transfer2.5~\cite{cosmostransfer} and Wan-Fun-Control~\cite{wanfuncontrol} across time steps and views. \algabbr{} produces realistic video quality while preserving object consistency across views. See Appendix~\ref{app:stacked} for more visual results.}
    \label{fig:visual_comparison}
    \vspace{-0.5em}
\end{figure*}

\subsection{Robot Simulation Data Generation}\label{ssec:sim_data}
We generate simulation trajectories using a pipeline built on MolmoSpaces~\cite{kim2026molmospaceslargescaleopenecosystem} and CuRobo~\cite{sundaralingam2023curobo}. For each episode, we instantiate a predefined simulation scene with target objects, a robot, and synchronized multi-view cameras. A scripted task-and-motion planner executes the task while recording multi-view RGB-D observations, proprioception, actions, and robot segmentation masks.
The planner uses offline grasp libraries for rigid and articulated objects, with precomputed antipodal end-effector poses for a parallel-jaw gripper. Grasps are selected by ranking~\cite{kim2026molmospaceslargescaleopenecosystem}, collision filtering, and IK feasibility checks. Once a feasible grasp is found, the planner executes task-specific primitives: pregrasp--grasp--lift--place--retreat for pick-and-place, constrained handle motion for articulated objects following an arc for hinge joints or a line for prismatic joints, and navigation--pregrasp--grasp--lift for mobile manipulation. If execution fails, the planner replans with another grasp. Episodes are retained only if object placement succeeds or the target joint displacement exceeds a task-defined threshold; otherwise, they terminate after repeated failures or a 300-step timeout. Appendix~\ref{app:simulation} provides additional planner and trajectory-generation details.

\subsection{\algabbr{} Data Generation}\label{ssec:method}
We use Claude Opus 4.6~\cite{anthropic2026opus46} to generate task-specific scene prompts. Together with the simulated depth videos and robot RGB-mask videos from~\cref{ssec:sim_data}, these prompts condition our video model to synthesize multi-view photorealistic RGB videos with prompt-controlled textures, backgrounds, and scene semantics. The robot RGB masks provide robot appearance and motion cues. Since the robot appearance is relatively simple and has a small sim-to-real gap, we directly use the simulator-rendered robot RGB masks as input.
Our video generation model can also insert distractor objects not present in simulation. We sample temporally consistent rectangular crops on the depth videos across views and augment the prompts with distractor descriptions, allowing the model to synthesize distractors in the cropped regions while preserving the original scene geometry and robot actions (\cref{fig:teaser}; Appendix~\ref{app:distractor}). Finally, pairing the generated RGB videos with simulator-provided actions and states yields realistic RGB-action-state trajectories for policy training.

\section{Video Generation Experiments}\label{sec:video_exp}

\begin{table}[t]
    \centering
        \caption{Quantitative results on generated video quality.}
        \label{tab:real_compare}
        \footnotesize
        \setlength{\tabcolsep}{3pt}
        \begin{tabular}{
            l
            S[table-format=2.1]
            S[table-format=3.1]
            S[table-format=2.2]
            S[table-format=1.3]
            S[table-format=1.3]
        }
            \toprule
            & \multicolumn{2}{c}{First-chunk} & \multicolumn{3}{c}{AR continuation} \\
            \cmidrule(lr){2-3} \cmidrule(lr){4-6}
            {Model} & {FID $\downarrow$} & {FVD $\downarrow$} & {PSNR $\uparrow$} & {SSIM $\uparrow$} & {LPIPS $\downarrow$} \\
            \midrule
            Wan-Fun-Control~\cite{wanfuncontrol}    & 62.5 & 267.8 & 20.30 & 0.738 & 0.175 \\
            Cosmos-Transfer2.5~\cite{cosmostransfer} & 42.8 & 164.5 & 13.26 & 0.481 & 0.423 \\
            RoboTransfer~\cite{liu2026robotransfercontrollablegeometryconsistentvideo} & 52.8 & 275.4 & 13.75 & 0.509 &0.418 \\
            \textbf{Ours}      & \bfseries 25.7 & \bfseries 93.9 & \bfseries 22.28 & \bfseries 0.784 & \bfseries 0.142 \\
            \bottomrule
        \end{tabular}
    \vspace{10pt}
        \caption{2AFC human study of the favor rate of \algabbr{} over baselines.}
        \label{tab:human_eval}
        \footnotesize
        \begin{tabular}{lcc}
            \toprule
            & \makecell{Multi-view\\consist.} & \makecell{Gen.\\quality} \\
            \midrule
            over Wan-Fun-Control~\cite{wanfuncontrol}    & 76.9\% & 90.6\% \\
            over Cosmos-Transfer2.5~\cite{cosmostransfer} & 74.9\% & 68.9\% \\
            over RoboTransfer~\cite{liu2026robotransfercontrollablegeometryconsistentvideo} & 85.7\% & 94.6\% \\
            \bottomrule
        \end{tabular}
        \vspace{-5pt}
\end{table}

\myparagraph{Setup.} We evaluate our video model against three baselines: Wan2.1-Fun-1.3B-Control~\cite{wanfuncontrol} (Wan-Fun-Control), a depth-conditioned variant of our base model Wan2.1-T2V-1.3B~\cite{wan}, Cosmos-Transfer2.5-2B~\cite{cosmostransfer} (Cosmos-Transfer2.5), the latest depth-control variant in the Cosmos foundation model family~\cite{cosmos} and RoboTransfer~\cite{liu2026robotransfercontrollablegeometryconsistentvideo}, a video generation model conditioned on depth, surface normals, and appearance reference images. We report FID~\cite{fid} and FVD~\cite{skorokhodov2022fvd} as distribution-level video quality metrics, and additionally report PSNR, SSIM~\cite{ssim}, and LPIPS~\cite{lpips} for paired evaluation against ground truth. We further conduct a two-alternative forced-choice (2AFC) human study as a complementary assessment. We recruited 50 raters, each comparing 20 side-by-side pairs of \algabbr{} and baseline videos in randomized left-right order and judging realism and multi-view consistency.

\myparagraph{Comparison with Baselines.} 
We evaluate generation quality on 500 held-out clips from DROID~\cite{droid} and AgiBot~\cite{agibot} under two inference-matched settings: \emph{first-chunk generation}, which synthesizes the initial chunk without previous-frame conditioning, and \emph{autoregressive continuation}, conditioned on a ground-truth previous frame. Wan-Fun-Control, RoboTransfer, and our model run at $416{\times}240$. Since Cosmos-Transfer2.5 fails at this resolution, outside its pretraining range, we run it at 480p and downsample outputs to 240p, favoring Cosmos. For RoboTransfer, we extract depth and normal conditions following its recipe.

As shown in~\cref{tab:real_compare}, \algabbr{} outperforms all baselines across all metrics and settings, despite Cosmos-Transfer2.5 using higher-resolution 480p inputs. Qualitative first-chunk results in~\cref{fig:visual_comparison} show that Wan-Fun-Control and Cosmos-Transfer2.5 lack multi-view consistency, causing shared objects such as the table to vary across views. Wan-Fun-Control also produces unrealistic robot frames because it is a general depth-to-video model without robot-specific adaptation. By repurposing a general video model with robot-specific data, \algabbr{} achieves both multi-view consistency and robot realism. For sim-depth inputs, where no ground-truth RGB reference exists, we use a 2AFC human study in~\cref{tab:human_eval}, \algabbr{} is preferred over all baselines in sim-to-real generation quality.

\begin{figure}[t]
    \centering
    \includegraphics[width=\linewidth]{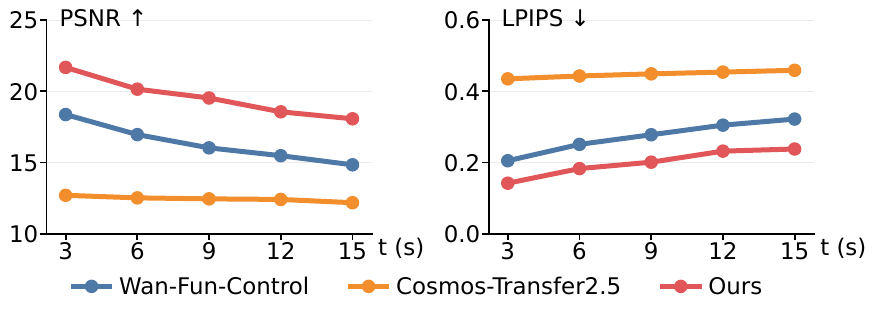}
    \caption{Per-frame PSNR and LPIPS along 15-second autoregressive rollouts, sampled every 3 seconds.}
    \label{fig:ar_decay}
    \vspace{-5pt}
\end{figure}
\myparagraph{Multi-chunk AR rollouts.} To assess autoregressive generation beyond a single chunk, we roll out Wan, Cosmos and \algabbr{} for 15\,s on 200 long real clips, initializing from a ground-truth previous frame and thereafter using its own output as the prev-frame anchor. As shown in \cref{fig:ar_decay}, our model leads at every time step, staying well ahead of both baselines in PSNR and LPIPS throughout. This indicates that \algabbr{} sustains generation quality across repeated autoregressive steps, enabling the multi-chunk rollouts needed to render full simulation trajectories into realistic videos.

\myparagraph{Multi-view consistency.}
We conduct a human study in which raters assess cross-view consistency across all three views. \algabbr{} is preferred over all baselines~\cref{tab:human_eval}, suggesting \algabbr{} can generate cross-view consistent videos. Appendix~\ref{app:stacked} reports additional qualitative comparisons and results for 3-View Stacked baselines.\looseness=-1

\myparagraph{Accelerated Inference.} To support large-scale data generation, we distill our controllable video model with DMD2~\cite{dmd2}, reducing the 50-step bidirectional diffusion sampler to 6 steps. The distillation runs for 2,000 iterations on a single H200 node with the standard DMD recipe. The distilled video model generates one 3-view clip in 4.29s on a single H200. See Appendix~\ref{app:acceleration} for acceleration details and TeaCache quality comparisons.

\begin{figure*}[t]
    \centering
    \includegraphics[width=\linewidth]{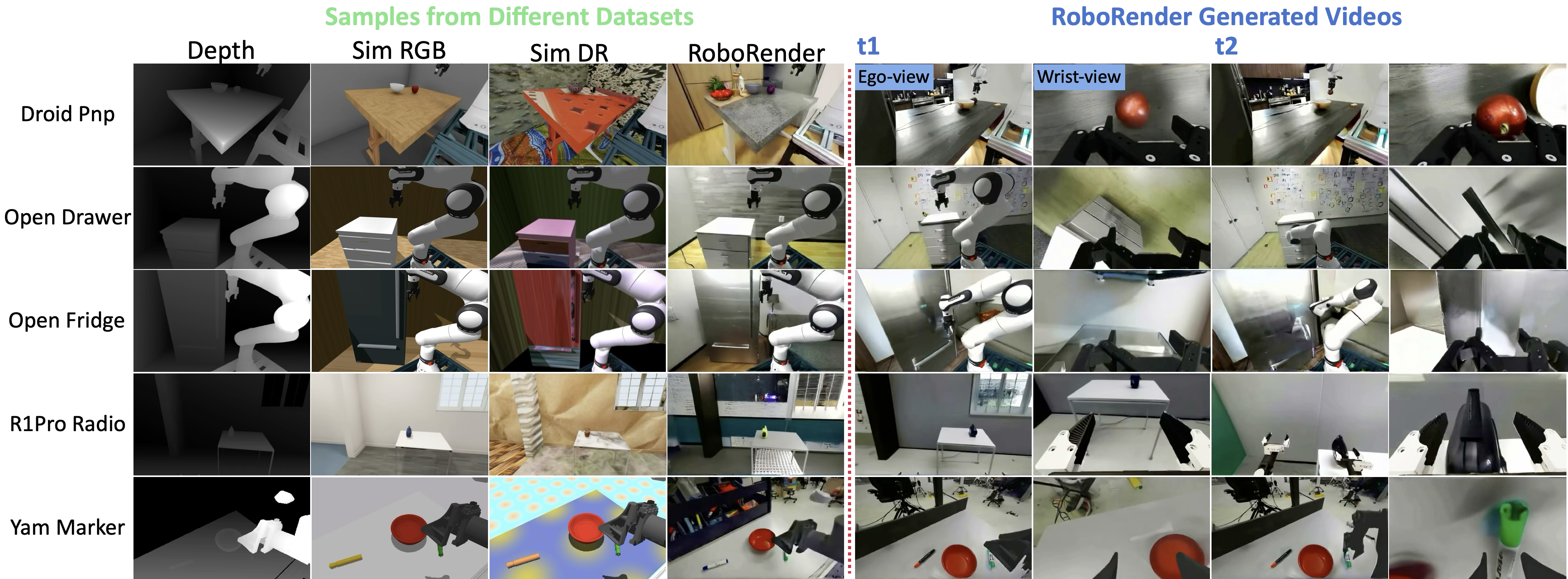}
    \caption{Dataset samples from different methods (Left) and \algabbr{} generated videos examples (Right).}
    \label{fig:data}
    \vspace{-5pt}
\end{figure*}

\section{Robot Physical Experiments}

\begin{figure}[t]
    \centering
    \includegraphics[width=\linewidth]{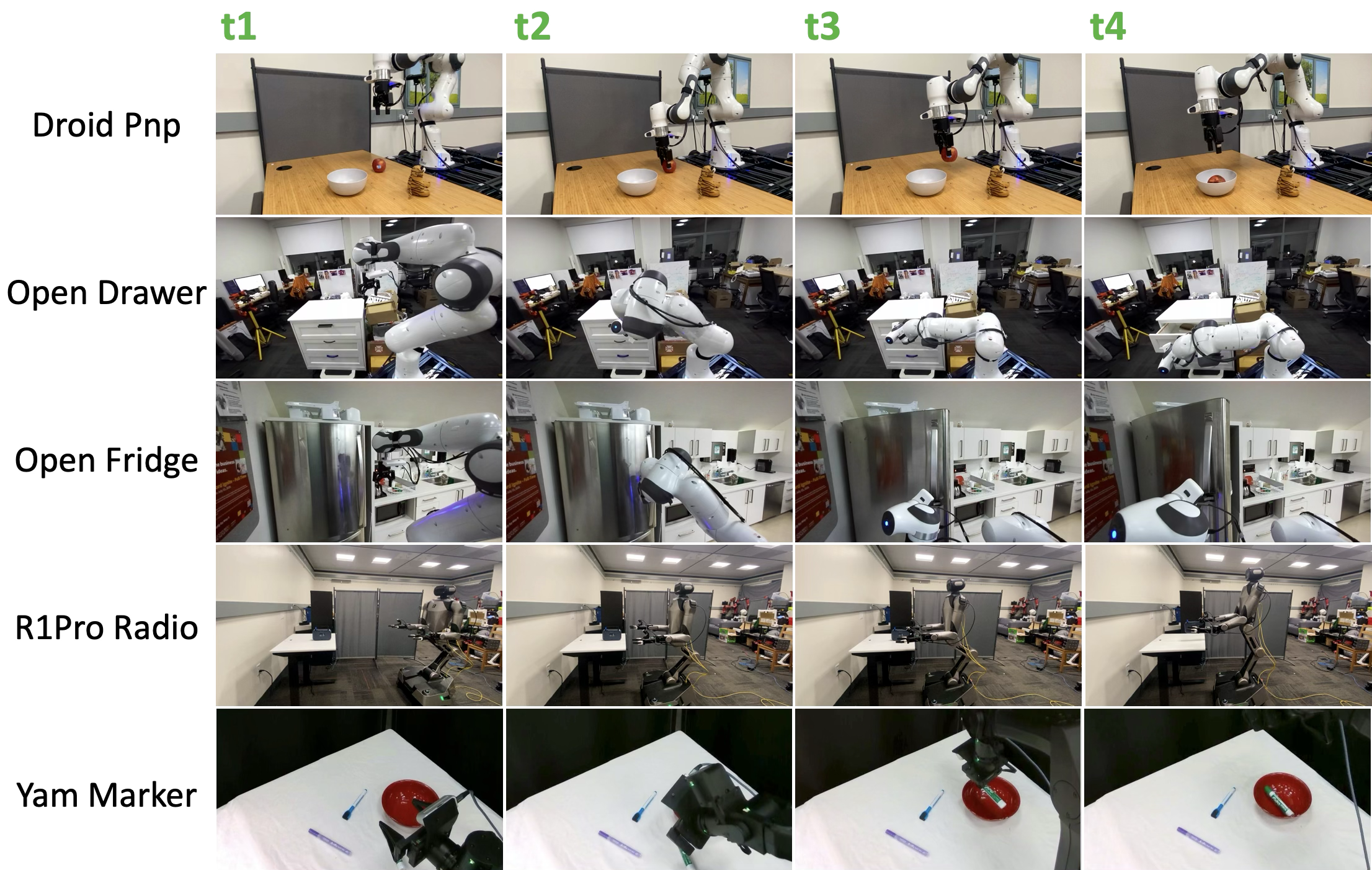}
    \caption{Robot physical rollout examples.}
    \label{fig:phy-rollout}
    \vspace{-15pt}
\end{figure}

We evaluate \algabbr{} on five pick-and-place (PnP) tasks, two articulated-object opening tasks (opening the fridge and drawer), and one mobile manipulation task in which the robot moves to a table and picks up a radio, as shown in~\cref{fig:phy-rollout}. For the four DROID PnP and opening tasks, we use a Franka robot with a Robotiq gripper, equipped with an external camera and a wrist-mounted camera following a DROID-style setup~\cite{droid}. For the mobile manipulation task, we use an R1Pro robot with one egocentric camera and two wrist cameras. We additionally perform one PnP task using the YAM Arm with one egocentric camera and one wrist camera, where the robot picks up a green marker from three markers and places in the red bowl. The video model has only been trained on robot data from Droid and Agibot, as in~\ref{ssec:video_gen}. Both R1Pro and YAM are unseen embodiment during the video model training.\looseness-1

We use different ways to randomize the configurations according to the type of task. For pick and place, we randomize the positions of task relevant objects and distractors, but for each trial, the positions are the same across different policies to ensure fair comparison. For the opening tasks (drawer and fridge) the randomization are done by changing the initial position and orientation of the robot. In the pick up the radio task, we randomize the physical setups by changing the mobile robot's initial position and radio's position on the table. See Appendix~\ref{app:physical_setup} for setup images.\looseness-1

\subsection{Policy Training}
\myparagraph{{Dataset Generation.}}
For the PnP and mobile manipulation tasks, we construct scenes with a table and the task relevant objects. We generate 1K trajectories per PnP task and 2K for mobile manipulation. For opening tasks, we create empty-room scenes with a single articulated object, generating 2K drawer-opening and 1K fridge-opening trajectories. To test cross-simulator applicability, DROID PnP and mobile manipulation data are generated in OmniGibson~\cite{li2022behavior}, while opening-task and YAM task data are generated in MuJoCo~\cite{todorov2012mujoco}. We randomize object size, table height, robot pose, and external camera pose, yielding the simulation dataset $\mathcal{D}_{sim}$ of RGB observations, robot states, and actions.
We construct a visual domain randomization baseline $\mathcal{D}_{dr}$ by replaying each simulation trajectory while randomizing object textures with DTD textures~\cite{cimpoi14describing} and lighting conditions. 
With the video model from~\cref{sec:video_exp} and following the procedure in~\cref{ssec:method}, we synthesize photorealistic RGB videos from the simulation data to form $\mathcal{D}_{gen}$. Examples of each datasets are shown in ~\cref{fig:data} and Appendix~\ref{app:policy_data}. For DROID PnP tasks, we additionally insert distractors using cropped depth patches: 2--5 patches in the external view, covering 20\%--50\% of the image, and 1--2 patches in the wrist view. 

\myparagraph{Training Setup.} We use $\pi_{0.5}$~\cite{intelligence2025pi05visionlanguageactionmodelopenworld} as the robot policy, which maps visual observations, language instructions, and robot proprioception to actions. We finetune the pretrained $\pi_{0.5}$ base model on each dataset to predict 32-step action chunks, training one joint policy for all PnP tasks and separate policies for each opening task and mobile manipulation. All policies are trained for 40k steps with batch size 72. At inference, we execute the first 16 predicted steps.

\begin{figure}[t]
    \centering
    \includegraphics[width=0.9\linewidth]{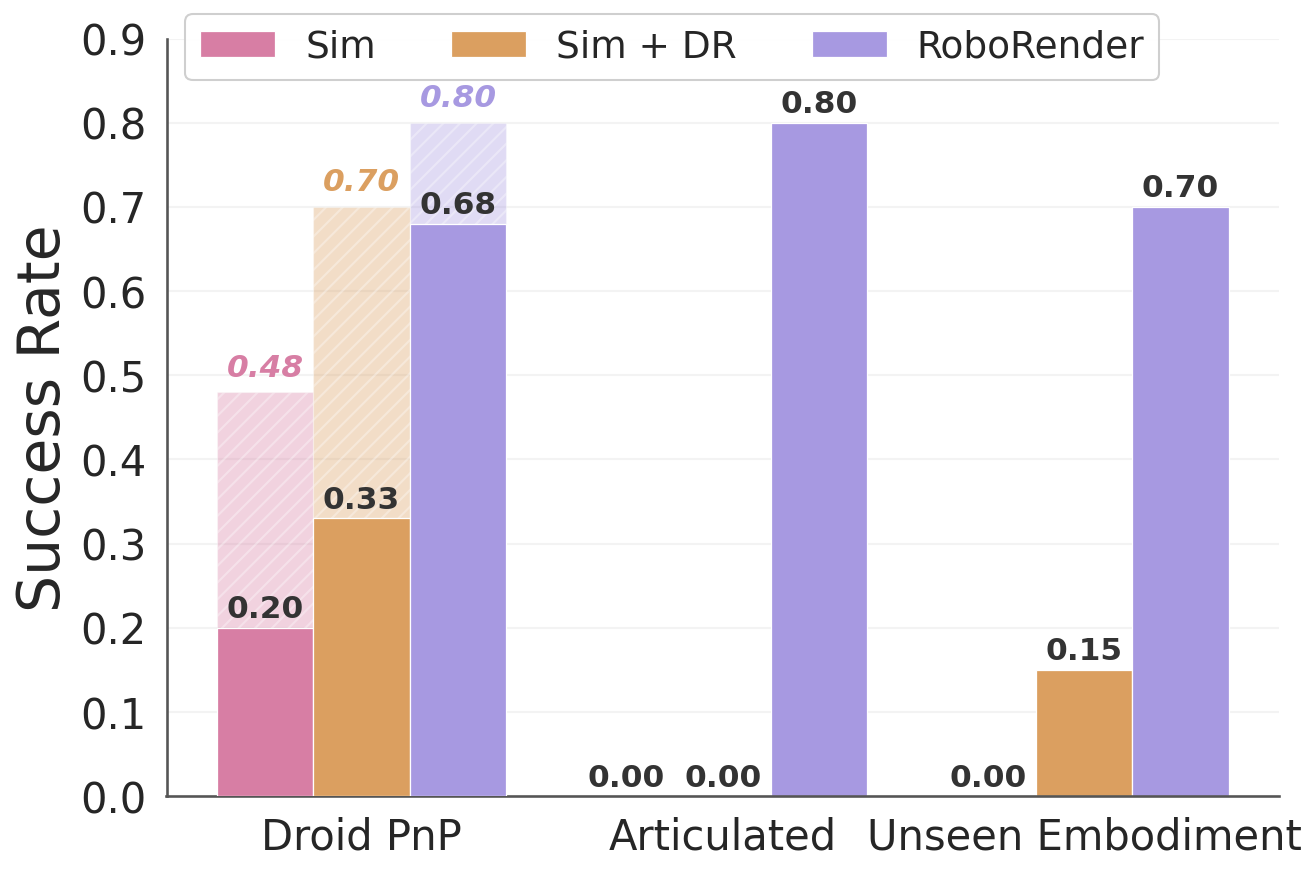}
    \caption{\textbf{Average success rate of each task category for each method.} For the Droid PnP category, the lighter \emph{hatched} extension on top of each bar shows the same
    method under the evaluation setup without distractor objects.}
    \label{fig:sr-bar}
    \vspace{-15pt}
\end{figure}

\subsection{Experiments and Analysis}
\begin{table*}[h!]
\centering
\caption{Real-world success rate across different tasks.}
\label{tab:real_exp}
\small
\resizebox{\textwidth}{!}{
\begin{tabular}{l ||c c c c |c c |c |c ||c }
\toprule
Embodiment & \multicolumn{6}{c|}{Droid} & R1Pro & Yam \\
\midrule
Data & Mug & Bowl & Apple & Egg & Open Fridge & Open Drawer & Pick up Radio & Place Green Marker& Avg. \\
\midrule
Sim        & 40\% & 30\% & 10\% & 0\%  & 0\% & 0\% & 0\% & 0\% & 10\% \\
Sim + DR   & 60\% & 20\% & 20\% & 30\%  & 0\%  & 0\% & 30\% & 0\% & 20\% \\
\algabbr{}  & \textbf{80\%} & \textbf{80\%} & \textbf{60\%} & \textbf{50\%}  & \textbf{70\%} & \textbf{90\%} & \textbf{70\%} & \textbf{70\%} & \textbf{71\%} \\
\bottomrule
\end{tabular}
}
\end{table*}
We compare \algabbr{} with two baselines: \textbf{Sim}, trained on $\mathcal{D}_{sim}$, and \textbf{Sim + DR}, trained on $\mathcal{D}_{dr}$. Each task is evaluated over 10 rollouts with varied robot starts and object locations. For PnP tasks, we add random table distractors to test robustness. All real-robot evaluations are conducted in a natural office environment with object size, geometry, camera viewpoint, and background differing from simulation (right column of~\cref{fig:phy-rollout}).~\cref{tab:real_exp} shows the results. \looseness=-1

\myparagraph{Does \algabbr{} mitigate the visual sim-to-real gap for zero-shot sim-to-real transfer?}
As shown in~\cref{tab:real_exp}, \algabbr{} achieves the highest average success rate across tasks, reaching 71\%, about 3.6$\times$ and 7.1$\times$ higher than Sim + DR and Sim, respectively. The category-level results in~\cref{fig:sr-bar} show the largest gains on opening tasks, where \algabbr{} is the only method with non-zero success. These tasks require recognizing fine visual details, such as drawer or fridge handles whose textures resemble the object body. Although visual domain randomization improves over raw simulation on PnP and mobile manipulation, it still fails on these visually precise tasks, suggesting that coarse appearance randomization is insufficient to close the visual sim-to-real gap. In contrast, \algabbr{} performs well across all task categories, showing that photorealistic RGB generation mitigates the visual sim-to-real gap for policy zero-shot sim-to-real transfer. 

We further test whether distractor insertion improves robustness to clutter. In the easier no-distractor setting (hatched bars in~\cref{fig:sr-bar}), Sim, Sim + DR, and \algabbr{} achieve success rates of 48\%, 70\%, and 80\%, respectively. When distractors are added during real-world evaluation, the baselines degrade substantially because their training data lacks clutter. By contrast, \algabbr{} shows only a small gap between the distractor and no-distractor settings, as it can insert random distractors during video generation. Thus \algabbr{} not only reduces the visual sim-to-real gap through photorealistic generation, but also improves robustness to cluttered real-world scenes by supporting realistic distractor insertion.

\myparagraph{Does \algabbr{} generalize to unseen robot embodiments?}
While trained only on DROID and AgiBot embodiments, the video model uses robot RGB-mask conditioning to generate videos for unseen robots. These masks are rendered in simulation, which suffices given the simple robot textures: on tasks with unseen R1Pro and YAM robots (\cref{fig:phy-rollout}), \algabbr{} substantially outperforms Sim and Sim + DR, suggesting that it can synthesize effective photorealistic training videos for new embodiments, reducing the visual sim-to-real gap.

\begin{figure}[t]
        \centering
        \includegraphics[width=\linewidth]{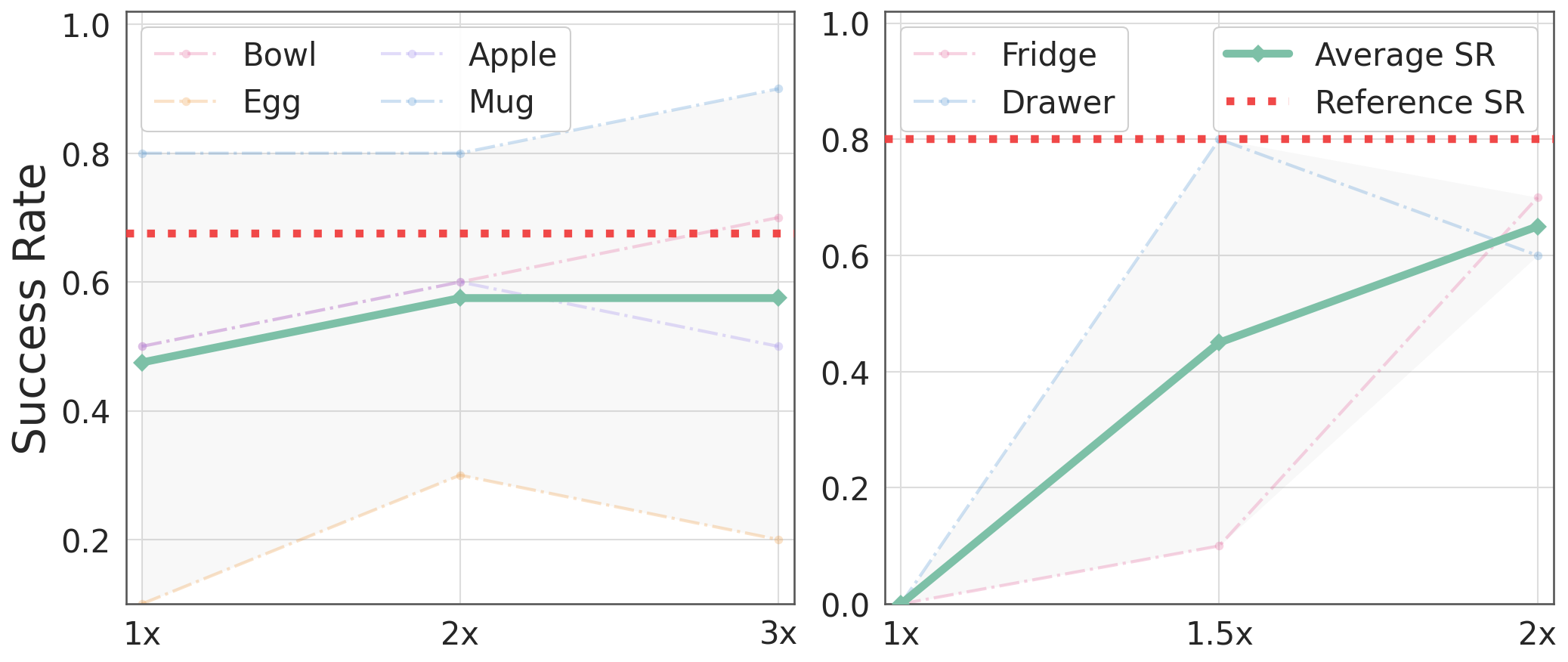}
        \caption{\textbf{Policy success rate v.s.\ the number of videos generated per unique simulation trajectory}, for PnP (left) and Opening (right).  \emph{Solid green} marks the per-category average, \emph{dash-dot} lines the individual tasks, and the \emph{dotted red} line the reference policy (one generated video per trajectory with 2x more simulation trajectores). Performance improves with more generated videos even though the underlying set of unique simulation trajectories is unchanged.}
        \label{fig:sr-scaling}
        \vspace{-5pt}
\end{figure}

\myparagraph{Does \algabbr{} outperform depth-conditioned policies?}
Depth-conditioned policies mitigate the visual sim-to-real gap but lack semantic information available in RGB observations. To demonstrate the value of high-quality RGB training videos, we compare \algabbr{} with depth-only policies trained on simulated depth for three PnP and mobile PnP tasks. At inference, the depth-only policies use depth estimated from RGB observations via VideoDepthAnything. \cref{tab:depth_baseline} shows that policies trained with \algabbr{} achieve substantially higher success rates across all three setups, suggesting that RGB semantic cues are important for identifying the correct object to complete the tasks.

\myparagraph{Does higher video generation quality lead to better policies?}
We compare policies trained on videos generated by \algabbr{} and baseline models on two opening tasks requiring high-quality RGB input. Specifically, we generated the same number of videos on the same simulation data using Cosmos and Wan. Since Cosmos and Wan don't take robot RGB mask videos, we add texture descriptions of the robot in the prompt for the baseline models. Results in~\cref{tab:video_baseline} show that Cosmos outperforms Wan but underperforms \algabbr{}. This aligns with the video quality results in~\cref{tab:real_compare}, suggesting that higher-quality RGB videos improve policy performance.

\begin{table}[t]
    \centering
    \caption{Comparison against policies trained with depth data.}
    \label{tab:depth_baseline}
    \begin{tabular}{lccc}
        \toprule
        Setup & YAM & Droid PnP & R1Pro \\
        \midrule
        Depth Only & 30\% & 25\% & 10\% \\
        \algabbr{} & \textbf{70\%} & \textbf{68\%} & \textbf{70\%} \\
        \bottomrule
    \end{tabular}
\end{table}

\begin{table}[t]
    \centering
    \caption{Comparison across video generation models for training data.}
    \label{tab:video_baseline}
    \begin{tabular}{lcc}
        \toprule
        Setup & Drawer & Fridge \\
        \midrule
        Wan-Fun-Control & 0\% & 0\% \\
        Cosmos-Transfer2.5 & 30\% & 0\% \\
        \algabbr{} & \textbf{90\%} & \textbf{70\%} \\
        \bottomrule
    \end{tabular}
\end{table}

\myparagraph{Does generating multiple videos per simulation trajectory improve policy learning?}
Because the video model is language-conditioned, it can generate diverse backgrounds, textures, and distractors from the same simulation trajectory. We study whether this visual diversity improves policy learning independent of collecting more unique trajectories.
We use only \textbf{half} of the simulation trajectories and vary the number of generated videos per trajectory by resampling language prompts and depth patches. For PnP tasks, we train with \(1\times\), \(2\times\), and \(3\times\) generations per trajectory using 500 trajectories per task. For Opening tasks, we train with \(1\times\), \(1.5\times\), and \(2\times\) generations per trajectory, using 500 fridge-opening and 1k drawer-opening trajectories. Results are shown in~\cref{fig:sr-scaling}, where the red dotted lines mark the performance of the policies in~\cref{tab:real_exp}, which train on \textbf{all} unique trajectories with one generated video each. While the gap to \(1\times\) reflects the value of physical trajectory diversity, additional generations recover much of this gap using only half the simulation data, especially on Opening tasks. On PnP tasks, average success rises from \(48\%\) at \(1\times\) to \(58\%\) at \(2\times\) and saturates at \(3\times\). On Opening tasks, average success increases more sharply, from \(0\%\) at \(1\times\) to \(45\%\) at \(1.5\times\) and \(65\%\) at \(2\times\), indicating that visual diversity is particularly useful for precise grounding, such as handle identification. These results suggest that \algabbr{} can generate diverse photorealistic videos from a fixed simulation dataset, improving policy learning efficiency without a proportional increase in simulation data.\looseness=-1

\section{Conclusion and Limitations}

We presented \algabbr{}, a framework that converts simulated trajectories into photorealistic RGB-action-state data with a robot-oriented video generation model. Conditioned on depth videos, language prompts, and robot RGB masks, \algabbr{} preserves simulator geometry and robot motion while synthesizing realistic textures, lighting, backgrounds, and distractors for policy training. Experiments show that policies trained on \algabbr{}-generated data achieve about \(7.1\times\) and \(3.6\times\) the average success rate of policies trained on raw simulation renderings and visual domain randomized data, respectively, across pick-and-place, articulated-object opening, and mobile manipulation tasks. These results demonstrate that \algabbr{} mitigates the visual sim-to-real gap for zero-shot transfer. Generating diverse videos from each simulated trajectory further improves policy learning efficiency without proportional increases in simulated data.

\myparagraph{Limitations.}
\algabbr{} has several limitations. First, it addresses the visual sim-to-real gap and therefore remains bounded by the simulator's fidelity in dynamics and contact modeling, making deformable-object tasks especially challenging. Second, although we show generalization to the unseen R1Pro embodiment, broader validation across robot morphologies, camera configurations, and real-world environments is needed. We also aim to scale \algabbr{} to more complex long-horizon tasks in the future.

\bibliographystyle{IEEEtran}
\bibliography{example}  %

\clearpage

\appendices
\makeatletter
\let\section\@IEEEappendixsavesection
\renewcommand{\thesection}{\Alph{section}}
\renewcommand{\thesectiondis}{\Alph{section}}
\makeatother
\section*{Appendix}

\renewcommand{\dbltopfraction}{0.85}
\renewcommand{\textfraction}{0.1}
\renewcommand{\dblfloatpagefraction}{0.75}
\setcounter{dbltopnumber}{3}
\setlength{\textfloatsep}{12pt plus 2pt minus 2pt}
\setlength{\dbltextfloatsep}{12pt plus 2pt minus 2pt}

\section{Video Model Training Data Curation}\label{app:data_curation}

\begin{figure*}[!t]
    \centering
    \includegraphics[width=\textwidth,height=0.75\textheight,keepaspectratio]{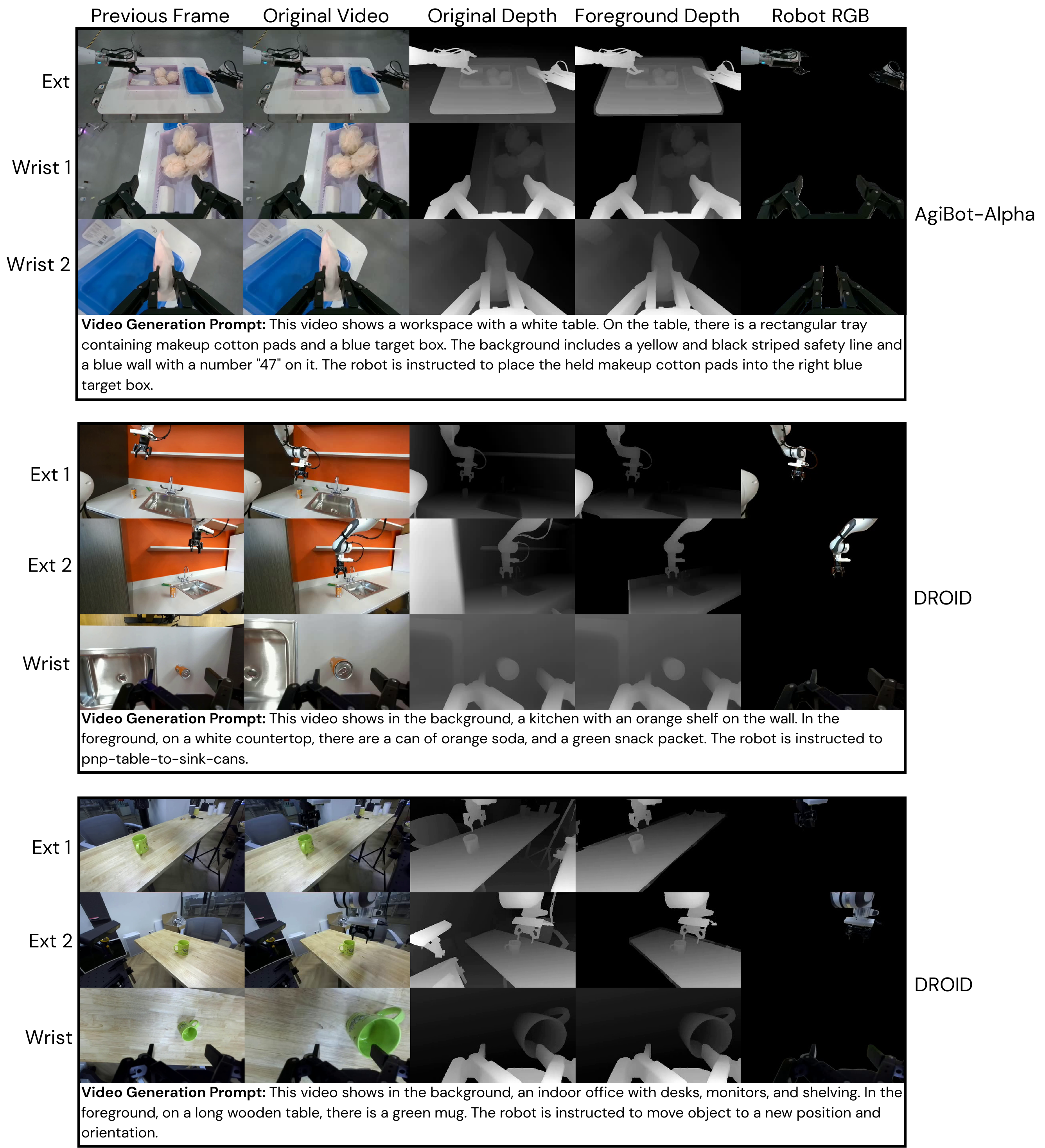}
    \caption{Example training samples for the video generation model, shown across camera views for AgiBot-Alpha and DROID. Each sample includes a Video Generation Prompt and five aligned conditions: Previous Frame, Original Video, Original Depth, Foreground Depth, and Robot RGB mask.}
    \label{fig:trainingdata}
    \vspace{-1em}
\end{figure*}

Training our video model requires large-scale paired (condition, video) data. However, most existing real-world robotics datasets~\cite{agibot, droid} are collected for policy learning rather than video generation, and lack the prompts, geometric conditions, and embodiment signals required by our model. We therefore design a scalable preprocessing pipeline that converts existing robot demonstrations into training-ready samples. Complete training data samples are shown in \cref{fig:trainingdata}.

For each demonstration clip, we extract a temporally consistent 81-frame window across all available camera views, and produce four condition signals aligned to this window: a scene description prompt, per-view depth videos, per-view robot-arm RGB and masks, and a pre-action frame.

\myparagraph{Scene Description Prompt.} We use a Vision-Language Model~\cite{qwenvl} in a two-pass scheme: a first pass produces a draft caption from the first frame, and a second pass refines it by cross-verifying objects across all non-wrist views. The model is constrained to output structured metadata covering background, work surface, and foreground objects, substantially reducing hallucination compared to single-pass captioning.

\myparagraph{Per-view Depth.} We extract per-view relative depth using Video-Depth-Anything~\cite{vda}, and additionally run FoundationStereo~\cite{foundationstereo} on samples with available stereo recordings. Both outputs are normalized via inverse-depth mapping. To avoid inductive bias from any single depth estimator, we randomly mix depth videos from different sources per view during training, improving robustness of the learned depth-to-appearance mapping.

\begin{strip}
    \begin{minipage}{\textwidth}
    \captionsetup{type=figure}
    \centering
    \begin{subfigure}{0.48\textwidth}
        \centering
        \includegraphics[width=\textwidth]{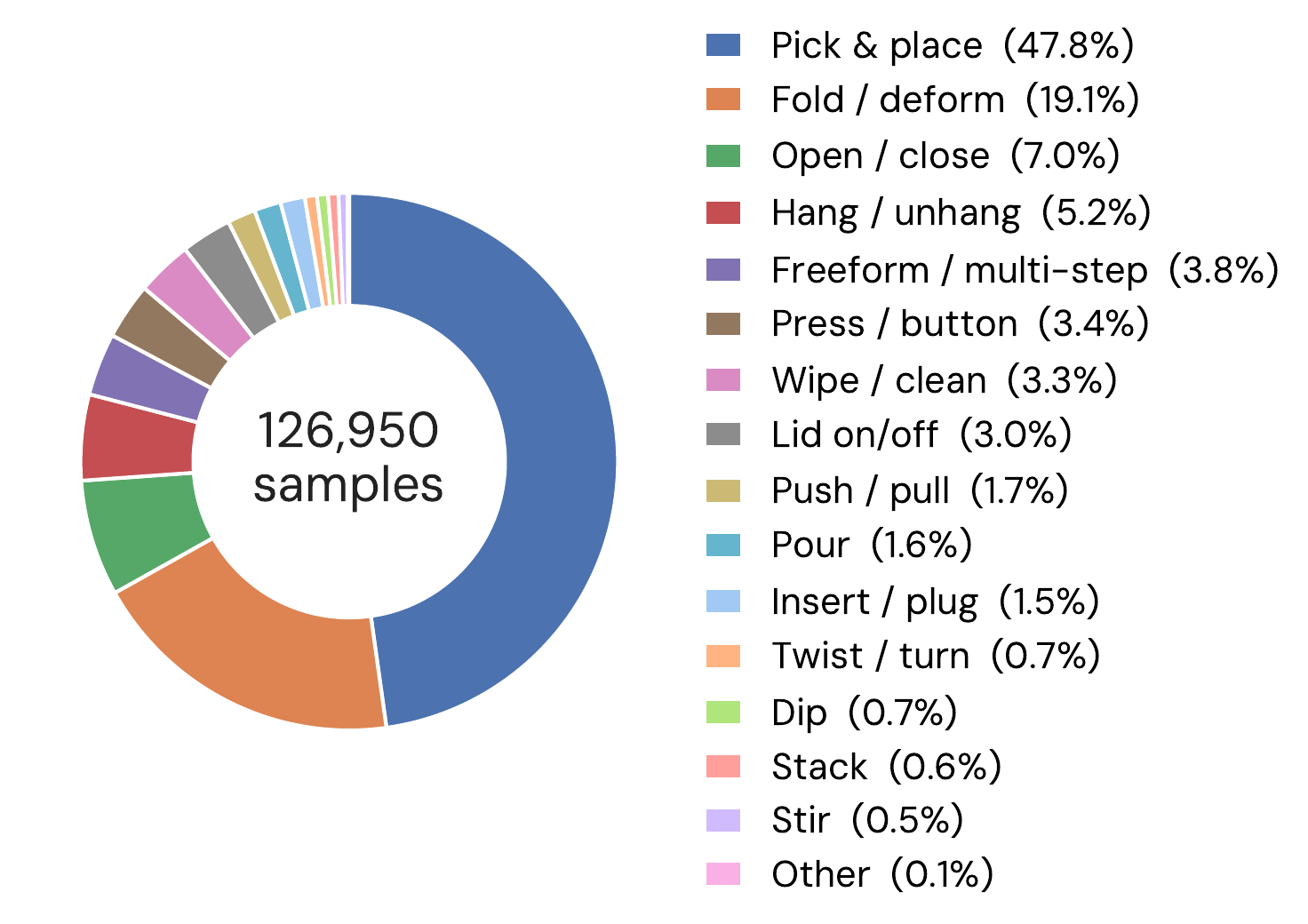}
        \caption{Pie chart showing the distribution of 16 unique tasks over
            the 126,950 training samples combining DROID and AgiBot-Alpha.}
        \label{fig:pie}
    \end{subfigure}
    \hfill
    \begin{subfigure}{0.48\textwidth}
        \centering
        \includegraphics[width=\textwidth]{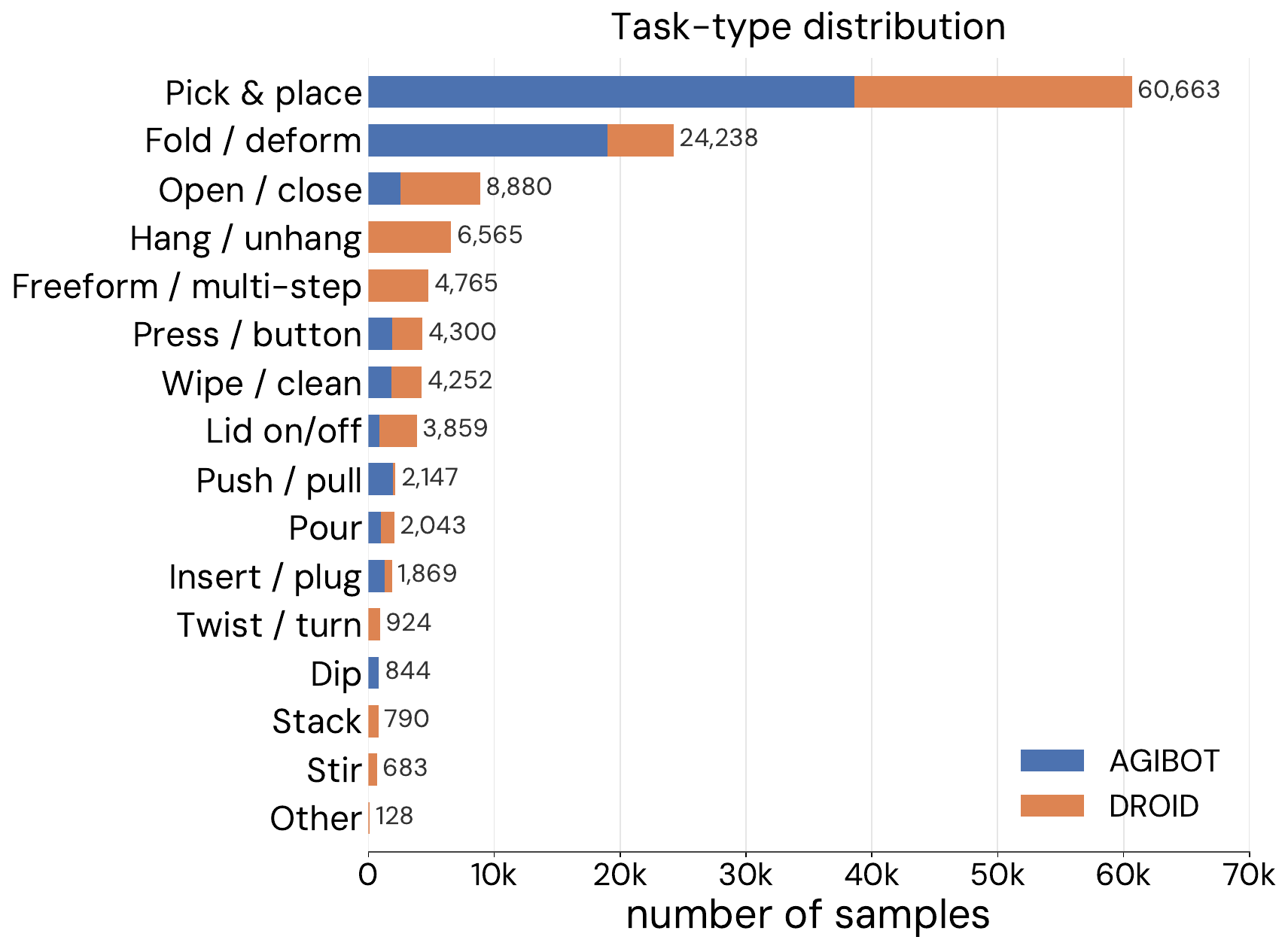}
        \caption{16 task categories breakdown with specific number of
            samples between DROID and AgiBot-Alpha.}
        \label{fig:hist}
    \end{subfigure}
    \caption{Training data composition for the video generation model.}
    \label{fig:training_composition}
    \end{minipage}
    \vspace{-1em}
\end{strip}

\myparagraph{Robot-arm RGB and Masks.} We use SAM3~\cite{sam3} to segment the robot arm and gripper across all views. These signals provide explicit embodiment grounding, isolating the agent's pose and motion from the rest of the scene.

\myparagraph{Pre-action Frame.} We include the frame immediately preceding the 81-frame clip as an additional condition, anchoring the visual appearance of the generated video and supporting autoregressive long-video generation at inference.

Our training dataset for the video model consists of ${\sim}130\mathrm{k}$ real-world robot manipulation samples drawn from two sources: ${\sim}70\mathrm{k}$ from AgiBot-World-Alpha~\cite{agibot} and ${\sim}60\mathrm{k}$ from DROID~\cite{droid}. We process all data at $416{\times}240$ resolution. We will release the processed paired data as an additional contribution to facilitate future research on robotics video generation. 

\cref{fig:hist}, \cref{fig:pie}, and \cref{tab:diversity} summarize the diversity of our training dataset. Our training dataset contains $126{,}950$ paired samples spanning $16$ manipulation task types. \cref{fig:pie} gives their overall proportions, led by pick \& place ($47.8\%$) and fold / deform ($19.1\%$), and the remaining $14$ types, such as open / close, hang / unhang, or stack, and \cref{fig:hist} shows per-type sample counts, with each bar split by source (AgiBot, $\sim\!70\mathrm{k}$; DROID, $\sim\!60\mathrm{k}$), and dataset-level totals are listed in \cref{tab:diversity}. In addition to showing the total sample number and task diversity, we also analyzed the number of distinct scenes in our training dataset. We found in total $39{,}692$ distinct scenes in our training datasets, which further demonstrates the high scene diversity in our training dataset.

\begin{table}[!ht]
\centering
\small
\setlength{\tabcolsep}{4pt}
\renewcommand{\arraystretch}{1.15}
\caption{Sample counts for our processed corpus, broken down by source dataset (AgiBot-World-Alpha and DROID) and combined. The dataset in total includes $126{,}950$ paired samples across $39{,}692$ distinct scenes and $16$ manipulation task types.}
\begin{tabular}{@{}lrrr@{}}
\toprule
\textbf{Dataset} & \textbf{\# Samples} & \makecell{\textbf{\# Distinct}\\\textbf{scenes}} & \makecell{\textbf{\# Task}\\\textbf{categories}} \\
\midrule
AGIBOT & 69,910 & 22,650 & 10 \\
DROID  & 57,040 & 17,042 & 15 \\
\midrule
\textbf{Total} & \textbf{126,950} & \textbf{39,692} & \textbf{16} \\
\bottomrule
\end{tabular}
\label{tab:diversity}
\end{table}

\begin{figure*}[!t]
    \begin{minipage}{\textwidth}
    \centering
    \includegraphics[width=0.49\textwidth]{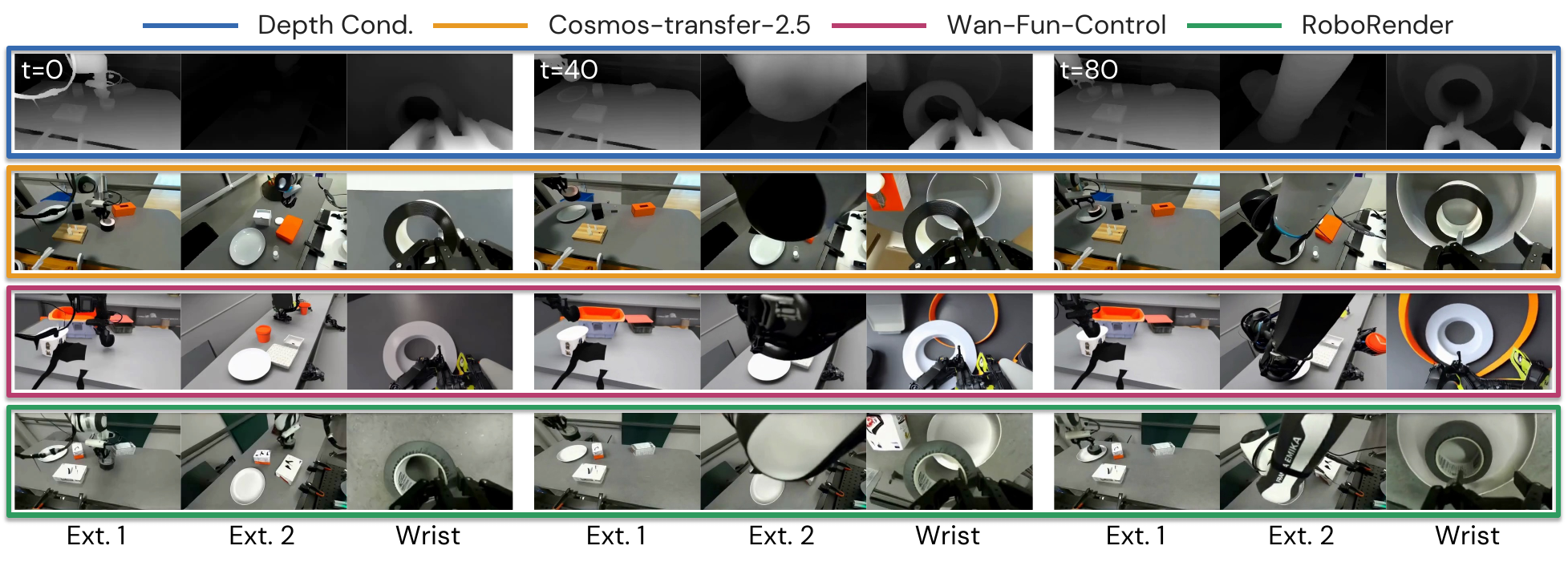}\hfill
    \includegraphics[width=0.49\textwidth]{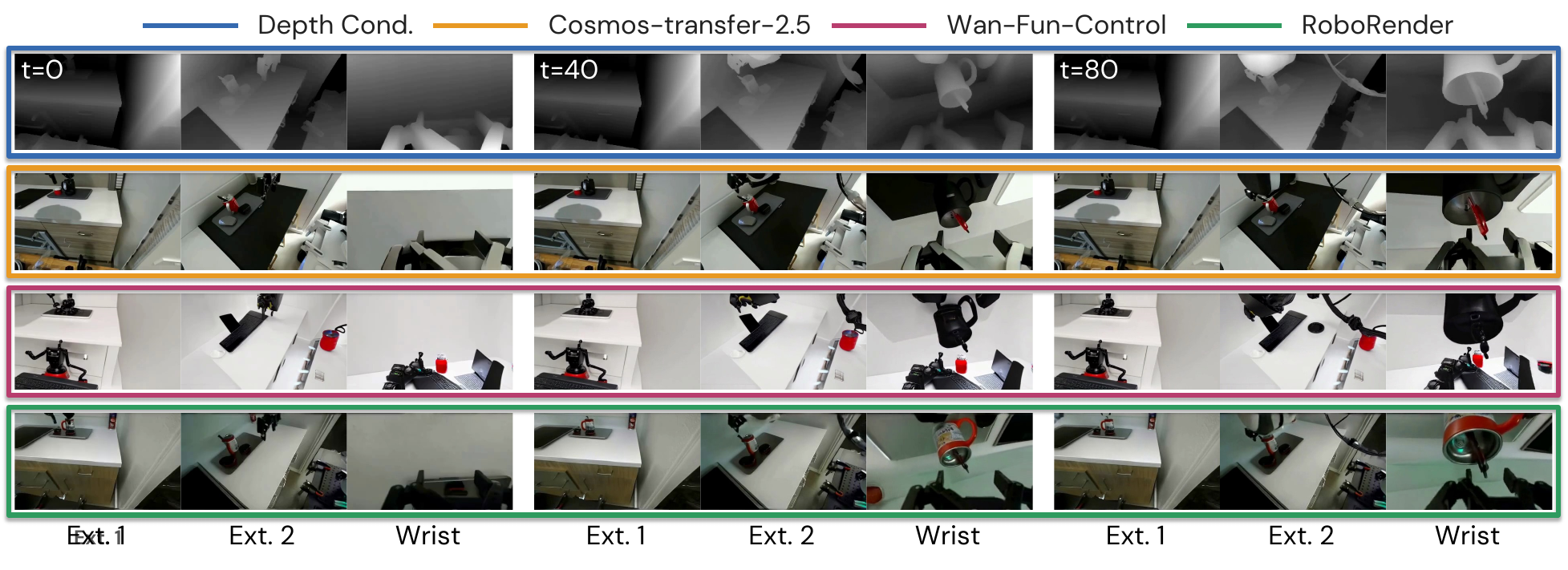}\\[3pt]
    \includegraphics[width=0.49\textwidth]{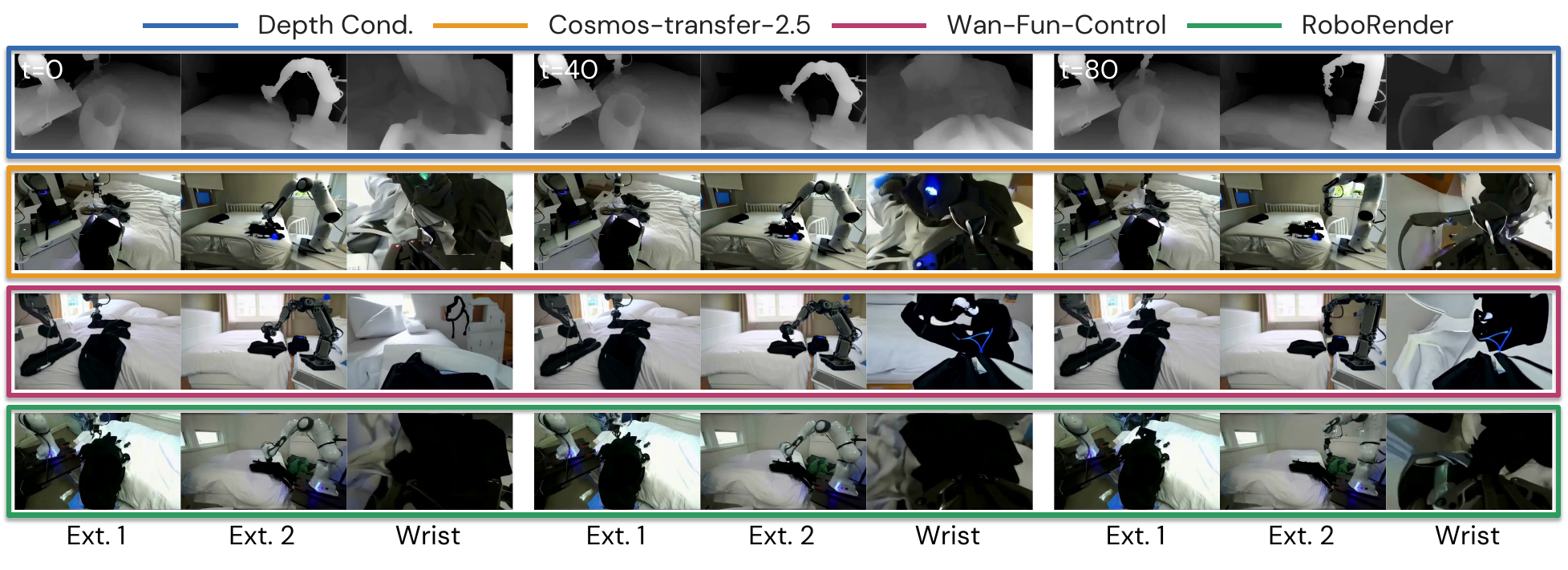}\hfill
    \includegraphics[width=0.49\textwidth]{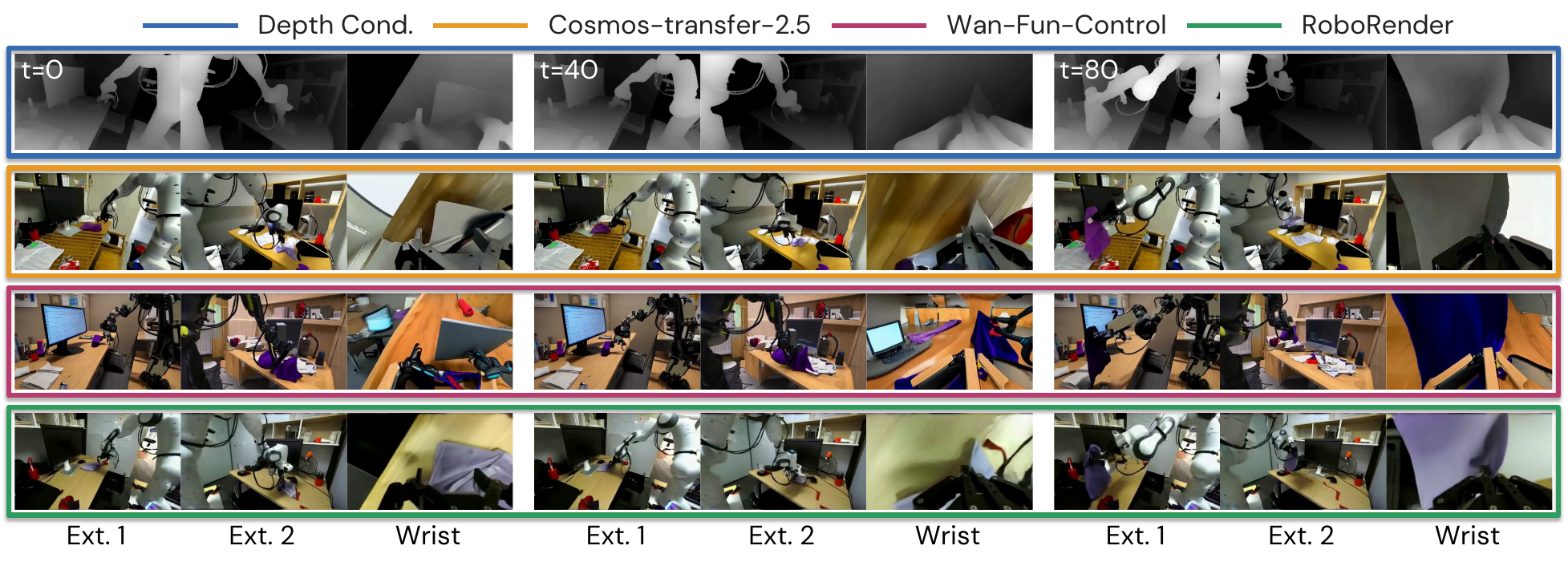}
    \caption{\textbf{Additional baseline comparison.} Conditioned on the same depth videos (top row), \algabbr{} renders realistic frames that stay consistent across the two external views and the wrist view, whereas Cosmos-Transfer2.5~\cite{cosmostransfer} and Wan-Fun-Control~\cite{wanfuncontrol} generate each view independently and produce cross-view inconsistencies.}
    \label{fig:more_baselines}
    \end{minipage}
\end{figure*}

\begin{figure*}[!t]
    \begin{minipage}{\textwidth}
    \centering
    \includegraphics[width=0.49\textwidth]{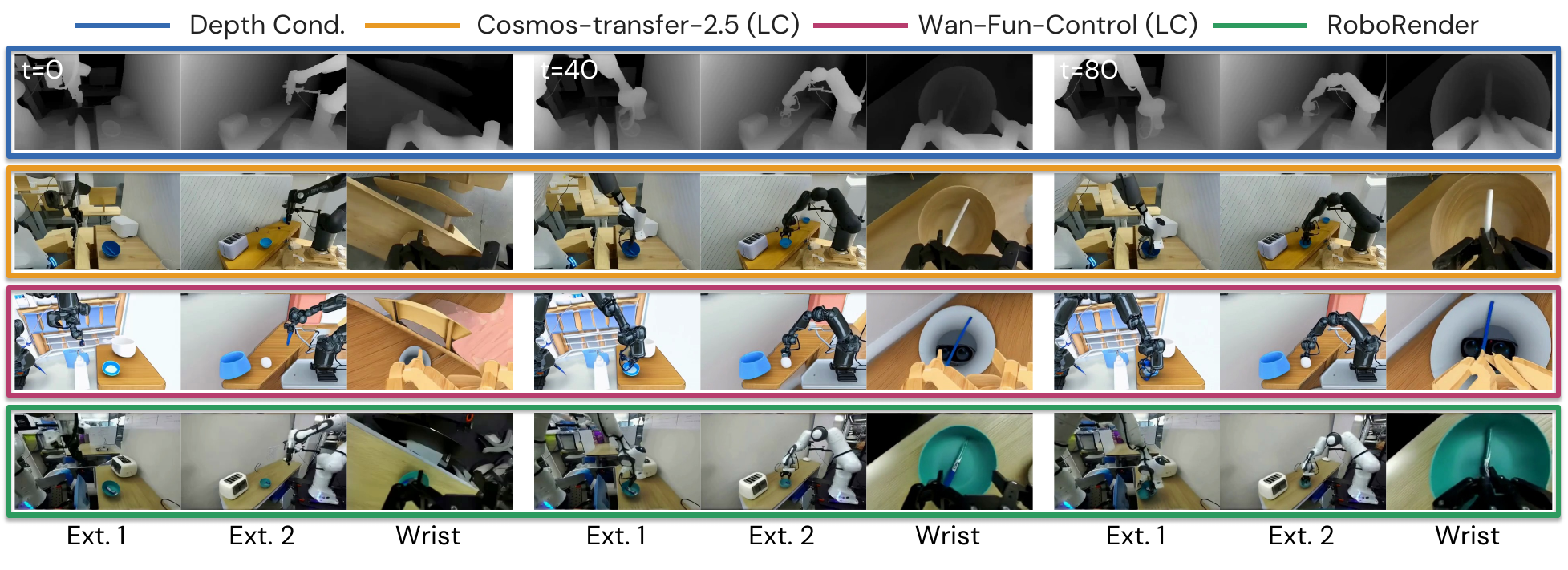}\hfill
    \includegraphics[width=0.49\textwidth]{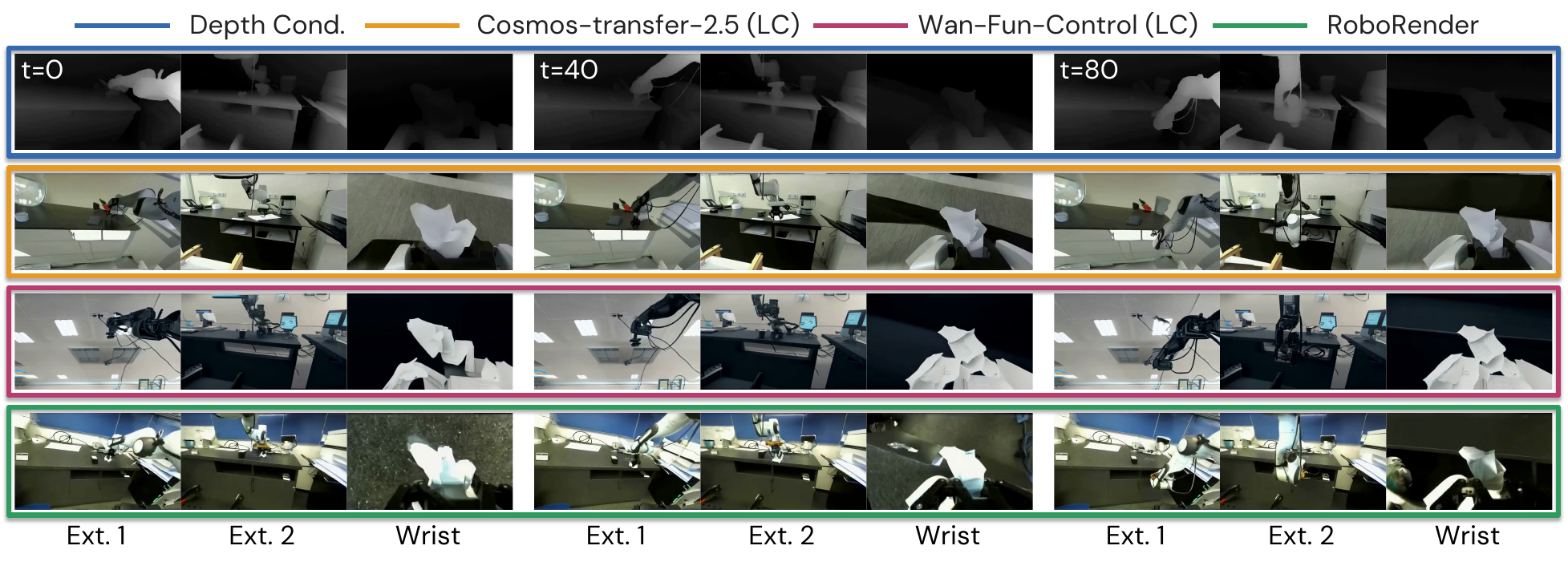}\\[3pt]
    \includegraphics[width=0.49\textwidth]{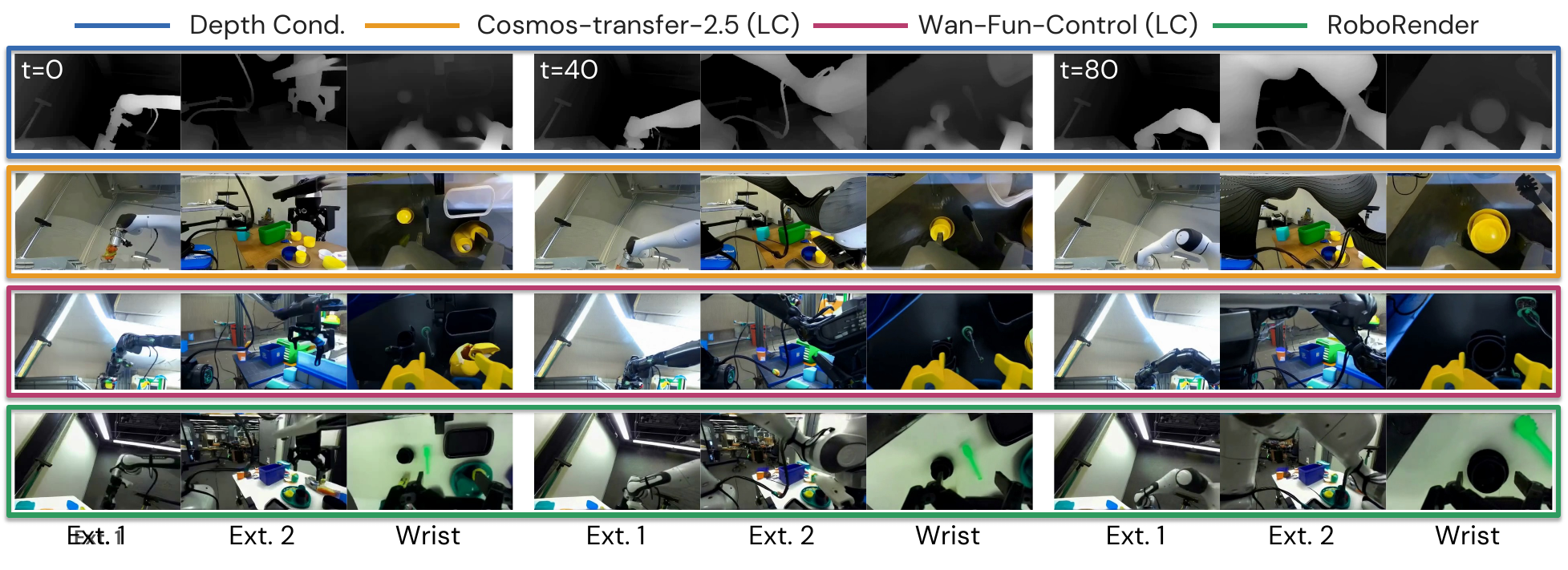}\hfill
    \includegraphics[width=0.49\textwidth]{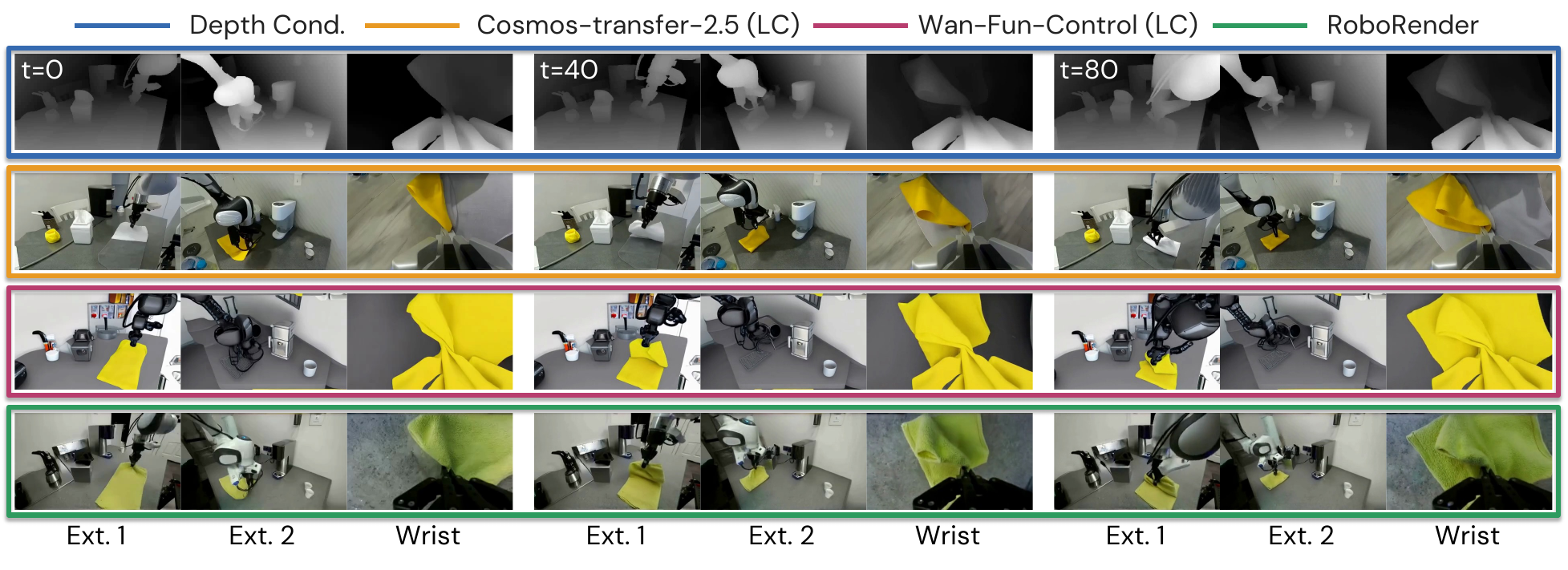}
    \caption{\textbf{Effect of 3-View Stacked.} Even with all three views stacked into a single input, Cosmos-Transfer2.5~\cite{cosmostransfer} (LC) and Wan-Fun-Control~\cite{wanfuncontrol} (LC) render shared objects, such as the table, inconsistently across views, whereas \algabbr{} stays consistent.}
    \label{fig:stacked_visual}
    \end{minipage}
\end{figure*}

\section{Video Generation Baseline Comparison}\label{app:stacked}

\myparagraph{Additional baseline comparisons.} \cref{fig:more_baselines} shows more qualitative comparisons against the baselines, where \algabbr{} produces realistic video quality while preserving object consistency across views.

\myparagraph{Effect of 3-View Stacked Input.} At inference, our model generates the three views jointly by concatenating them in the latent space. To confirm that our advantage comes from dedicated multi-view training rather than this inference scheme alone, we grant both baselines the same latent concatenation, feeding all three views as a single stacked input (the \emph{3-View Stacked} variants). As shown in \cref{tab:first_chunk_stacked}, latent concatenation does not close the gap: it only marginally improves Wan-Fun-Control~\cite{wanfuncontrol} and even degrades Cosmos-Transfer2.5~\cite{cosmostransfer}, while our model remains far ahead. \cref{fig:stacked_visual} shows the same qualitatively, where the stacked baselines still render shared objects inconsistently across views. These results show that latent concatenation at inference alone is insufficient for multi-view consistency, which instead requires dedicated multi-view training.

\begin{table}[!ht]
    \centering
    \caption{Effect of 3-View Stacked input on baselines (first-chunk synthesis, real input).}
    \label{tab:first_chunk_stacked}
    \small
    \setlength{\tabcolsep}{4pt}
    \begin{tabular}{@{}lcc@{}}
        \toprule
        Model & FID $\downarrow$ & FVD $\downarrow$ \\
        \midrule
        Wan-Fun-Control & 62.5 & 267.8 \\
        \makecell[l]{Wan-Fun-Control\\(3-View Stacked)} & 61.8 & 255.6 \\
        Cosmos-Transfer2.5 & 42.8 & 164.5 \\
        \makecell[l]{Cosmos-Transfer2.5\\(3-View Stacked)} & 57.1 & 236.0 \\
        Ours & \textbf{25.7} & \textbf{93.9} \\
        \bottomrule
    \end{tabular}
\end{table}

\begin{figure*}[!t]
    \begin{minipage}{\textwidth}
    \centering
    \includegraphics[width=\textwidth]{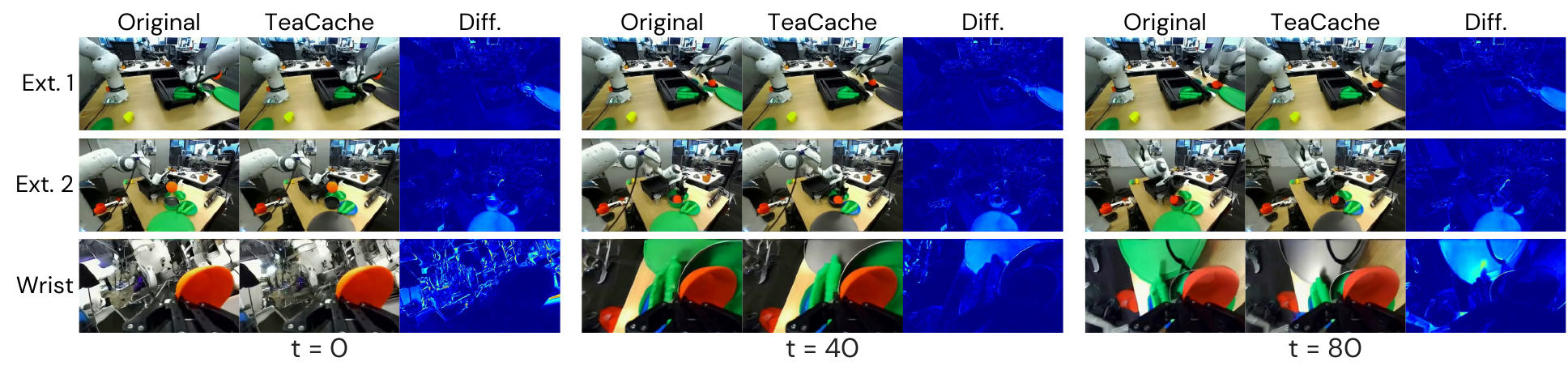}\\[3pt]
    \includegraphics[width=\textwidth]{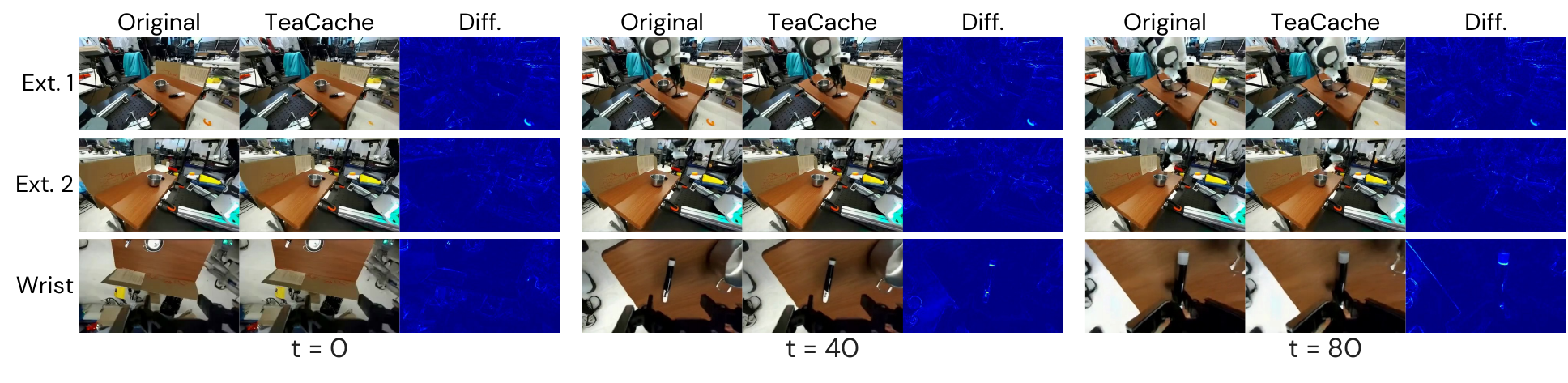}
    \caption{\textbf{Quality preservation under TeaCache.} For each view and time step, we show the default-pipeline frame (Original), the TeaCache frame, and their per-pixel difference (Diff). The difference is small and confined to fine texture, leaving scene layout, objects, and robot pose unchanged.}
    \label{fig:teacache_quality}
    \end{minipage}
\end{figure*}

\begin{figure*}[!t]
    \centering
    \begin{minipage}[t]{0.49\textwidth}
        \vspace{0pt}\centering
        \includegraphics[width=\linewidth,viewport=0 645 834 1305,clip]{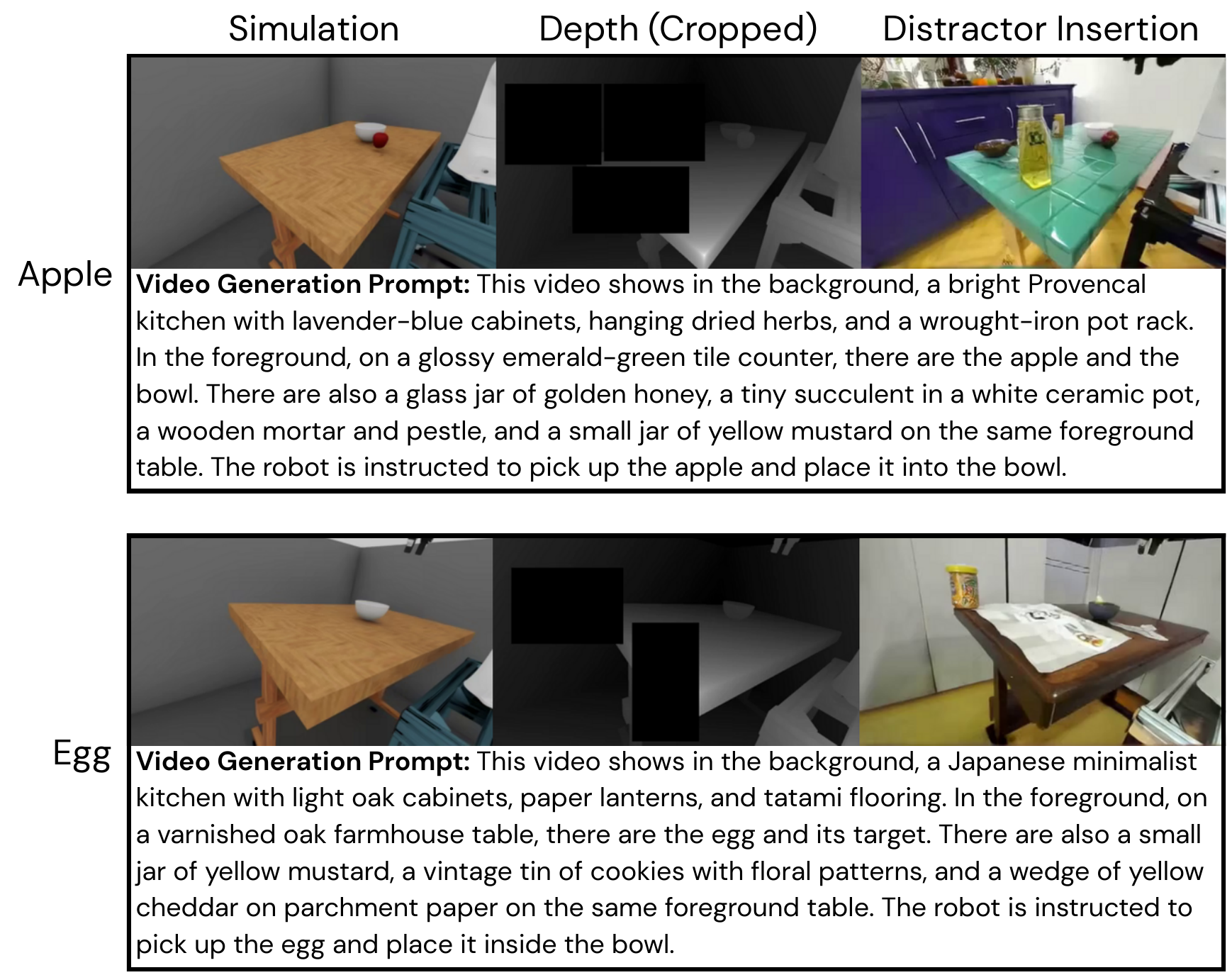}
    \end{minipage}\hfill
    \begin{minipage}[t]{0.49\textwidth}
        \vspace{0pt}\centering
        \includegraphics[width=\linewidth,viewport=0 1270 834 1305,clip]{figures/depth_distractors.pdf}\\
        \includegraphics[width=\linewidth,viewport=0 0 834 619,clip]{figures/depth_distractors.pdf}
    \end{minipage}
    \caption{Distractor insertion via depth cropping. We crop regions of the simulation depth (middle, black), eliminating those areas from geometric constraint so that RoboRender (right) synthesizes photorealistic distractor objects in these places while preserving the task-relevant geometry.}
    \label{fig:distractor}
\end{figure*}

\section{On Accelerating Video Model Generation}\label{app:acceleration}
Rendering an entire simulation dataset into video requires many diffusion forward passes, so inference cost is the main bottleneck for scaling up data generation. We explore two directions to reduce it.

\myparagraph{Accelerated Inference.} We accelerate inference with TeaCache~\cite{teacache}, which caches and reuses intermediate diffusion features to skip redundant computation across denoising steps. On a single NVIDIA H200 GPU with 50 denoising steps and a CFG scale of 5.0, our default pipeline generates a 3-view 81-frame clip in 68.8\,s, while the TeaCache variant (threshold $\tau=0.10$) produces the same clip in 35.3\,s, a 1.95$\times$ speedup. As shown in \cref{fig:teacache_quality}, the error map is small and the two pipelines are visually close in quality.

\myparagraph{Step distillation.} We further distill our controllable video model with DMD2~\cite{dmd2}, reducing the 50-step bidirectional diffusion sampler to 6 steps. The distillation runs for 2{,}000 iterations on a single node with the standard DMD recipe. The distilled model generates one clip in 4.29s on a single H200 GPU.

\section{Distractor Insertion}\label{app:distractor}

\cref{fig:distractor} illustrates how RoboRender enables controllable distractor
insertion. Starting from the simulation depth (middle), we randomly crop out regions of the depth map on the table (shown in black), removing the geometric constraint in those areas in both views. When
RoboRender renders the photorealistic video (right), these unconstrained regions are no longer pinned to the simulated geometry, allowing the model to synthesize plausible
distractor objects and background clutter in these places. The inserted distractors remain consistent over time, augmenting the visual diversity of the training data without introducing geometric artifacts in the task-relevant regions.

\begin{figure*}[!t]
    \includegraphics[width=1\textwidth]{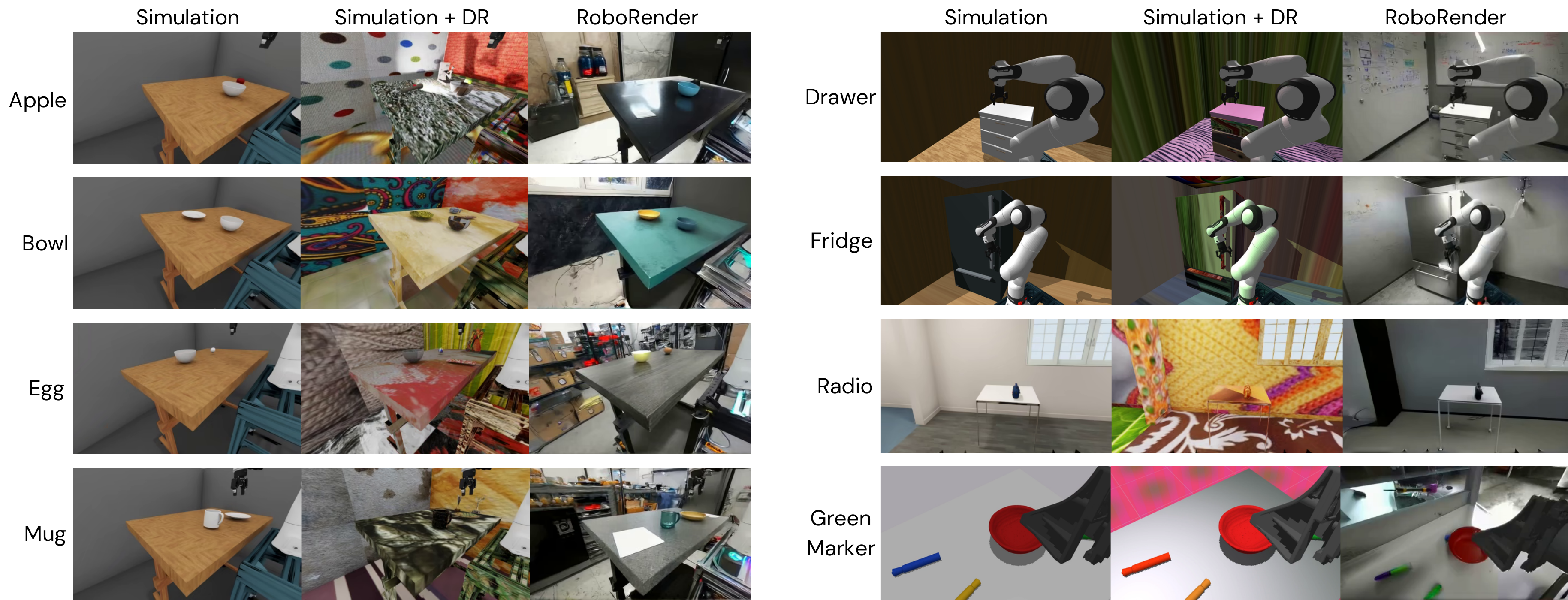}
    \caption{Visualization of policy training data across eight evaluation tasks between raw Simulation, Simulation + Domain Randomization, and \textbf{RoboRender} (ours). Comparing to the other two methods, RoboRender yields photorealistic textures and lighting while preserving geometry, minimizing the appearance gap to the real world.}
    \label{fig:datavis}

\end{figure*}

\begin{figure*}[!t]
    \begin{minipage}{\textwidth}
    \centering
    \includegraphics[width=\textwidth]{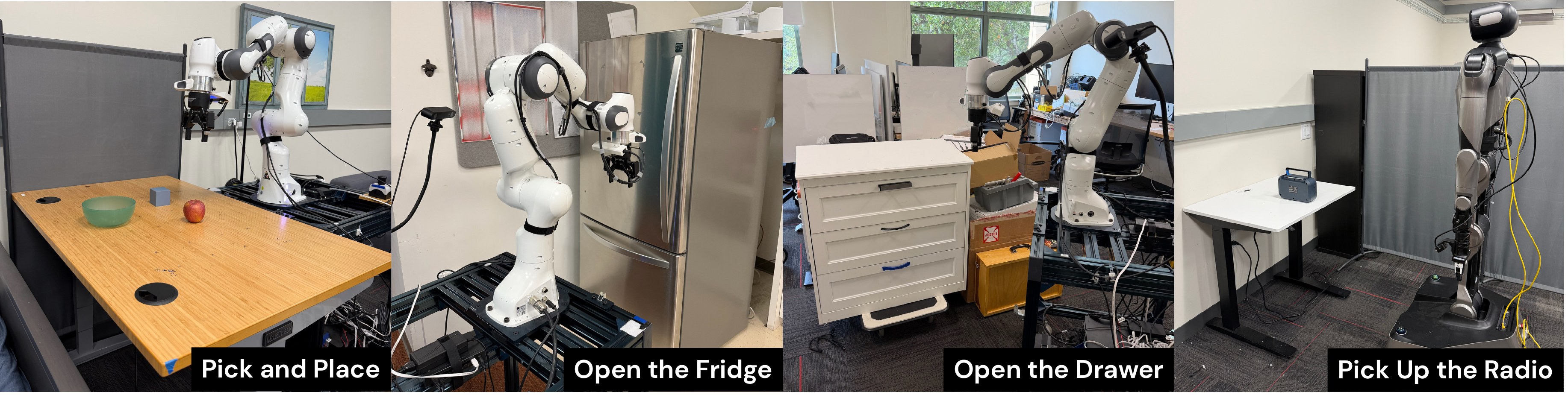}
    \caption{Physical setups and the initial poses of the robot in completion for each task}
    \label{fig:phy-setup}
    \end{minipage}
\end{figure*}

\section{Robot Simulation Data Generation}\label{app:simulation}
We generate task trajectories with a pipeline built on MolmoSpaces~\cite{kim2026molmospaceslargescaleopenecosystem} and CuRobo~\cite{sundaralingam2023curobo}. For each episode, we instantiate a scene with a target object or articulated object--joint pair, a robot at a feasible base pose, and synchronized multi-view cameras. A scripted task-and-motion planner executes the task while recording RGBD observations of each view, robot proprioception, and robot actions. We also record the robot segmentation mask videos from the simulation.

Actions are produced by a scripted planner with offline grasp libraries for both rigid-object and articulated-object manipulation. Each manipulable target is associated with precomputed antipodal end-effector poses for a parallel-jaw gripper; for articulated tasks, grasp libraries are defined per object--joint pair, such as a door or drawer handle. At episode start, candidate grasps are transformed into the world frame and augmented with 180-degree rotations to account for gripper symmetry.

The planner selects a grasp through ranking, collision filtering, and IK feasibility checking. Candidates are ranked by reachability, orientation alignment, approach stability, and proximity to the task-relevant manipulation region, such as the object center of mass for pick-and-place or the handle midpoint for articulated objects~\cite{kim2026molmospaceslargescaleopenecosystem}. The top candidates are collision-checked against the object, support surface, clutter, and robot links, then passed through batched IK checks. The first feasible grasp is selected.

After grasp selection, the planner executes task-specific motion primitives. Pick-and-place tasks follow a pregrasp--grasp--lift--place--retreat sequence, while articulated tasks grasp the handle and move along the target joint constraint, using an arc for hinge joints and a linear path for prismatic joints. For mobile manipulation, the planner follow navigation--pregrasp--grasp--lift sequence. If execution fails due to IK loss, motion stalling, downstream infeasibility, or gripper slip, the planner replans with a different grasp. Success is measured by object placement or object lifting or by joint displacement crossing a task-defined threshold. The episode fails after the maximum number of retries is exceeded or a timeout of 300 simulator steps. Successful trajectories are retained for policy training. \looseness=-1

\section{Policy Training Data Visualization}\label{app:policy_data}

\cref{fig:datavis} shows exemplar data we used to train each robot evaluation policy for completing the 8 tasks. Raw simulation (left) provides clean geometry but exhibits a large appearance gap from the real world, specifically synthetic textures, uniform lighting, and an unrealistic background. Standard domain randomization on simulation data (middle) perturbs textures and colors to widen the training distribution, but the resulting frames are visually incoherent and still fail to match real-world material and lighting statistics. Our method \textbf{RoboRender} (right) generates photorealistic observations that preserve the underlying scene geometry while producing realistic surface textures, reflections, and lighting, which closes the appearance gap without sacrificing the geometric grounding required for the task. This higher visual realism allows policies trained on RoboRender data to transfer zero-shot to the real robot across different manipulation tasks, including opening tasks and mobile robot tasks that are often considered to be more difficult.

\section{Physical Setup}\label{app:physical_setup}

\cref{fig:phy-setup} shows the physical setups and initial poses of our robots in completion for each task. Note that we used different ways to randomize the configurations according to the type of task. For pick and place, we randomized the positions of task relevant objects and distractors, but for each trial, the positions are the same across different policies to ensure fair comparison. For the opening tasks (drawer and fridge) the randomization was done by changing the initial position and orientation of the robot. In the pick up the radio task, we randomize the physical setups by changing the mobile robot's initial position and radio's position on the table.

\end{document}